\documentclass[runningheads]{llncs}

\usepackage{eccv}

\usepackage[dvipsnames]{xcolor}

\usepackage{amsmath,amsfonts,bm}

\def\eqref#1{equation~\ref{#1}}

\def\1{\bm{1}}

\DeclareMathAlphabet{\mathsfit}{\encodingdefault}{\sfdefault}{m}{sl}
\SetMathAlphabet{\mathsfit}{bold}{\encodingdefault}{\sfdefault}{bx}{n}

\usepackage{algorithm}
\usepackage{algorithmic}
\usepackage[utf8]{inputenc} 
\usepackage{booktabs}       
\usepackage{amsfonts}       
\usepackage{nicefrac}       
\usepackage{microtype}      
\usepackage{xcolor}         
\usepackage{dsfont}
\usepackage{multirow}
\usepackage{newfloat}
\usepackage{listings}
\usepackage{tikz}
\usepackage{eccvabbrv}
\usepackage{xcolor,colortbl}
\usepackage{graphicx}
\usepackage{booktabs}

\usepackage[accsupp]{axessibility}  

\makeatletter
\renewcommand{\fnum@figure}{Figure \thefigure}
\makeatother

\usepackage{pifont}
\newcommand{\cmark}{\ding{51}}%
\newcommand{\xmark}{\ding{55}}%

\usepackage{hyperref}

\usepackage{orcidlink}
\usepackage{wrapfig,booktabs}

\begin{document}

\title{    
    Revisiting Supervision for Continual Representation Learning
} 


\author{
Daniel~Marczak\thanks{Corresponding author, email: \href{mailto:daniel.marczak.dokt@pw.edu.pl}{daniel.marczak.dokt@pw.edu.pl}}\inst{1,2}\orcidlink{0000-0002-6352-9134} \and
Sebastian~Cygert\inst{1,3}\orcidlink{0000-0002-4763-8381} \and \\
Tomasz~Trzciński\inst{1,2,4}\orcidlink{0000-0002-1486-8906} \and
Bartłomiej~Twardowski\inst{1,5,6}\orcidlink{0000-0003-2117-8679}
}

\authorrunning{D.~Marczak et al.}

\institute{
IDEAS~NCBR \and
Warsaw~University~of~Technology \and 
Gdańsk~University~of~Technology \and 
Tooploox \and
Autonomous~University of~Barcelona \and 
Computer~Vision~Center
}

\maketitle

\begin{abstract}
In the field of continual learning, models are designed to learn tasks one after the other. While most research has centered on supervised continual learning, there is a growing interest in unsupervised continual learning, which makes use of the vast amounts of unlabeled data. Recent studies have highlighted the strengths of unsupervised methods, particularly self-supervised learning, in providing robust representations. The improved transferability of those representations built with self-supervised methods is often associated with the role played by the multi-layer perceptron projector. In this work, we depart from this observation and reexamine the role of supervision in continual representation learning. We reckon that additional information, such as human annotations, should not deteriorate the quality of representations. Our findings show that supervised models when enhanced with a multi-layer perceptron head, can outperform self-supervised models in continual representation learning. This highlights the importance of the multi-layer perceptron projector in shaping feature transferability across a sequence of tasks in continual learning. The code is available on \href{https://github.com/danielm1405/sl-vs-ssl-cl}{github}.
\end{abstract}

\section{Introduction}

\begin{figure}[t]
    \centering
    \includegraphics[width=0.7\textwidth]{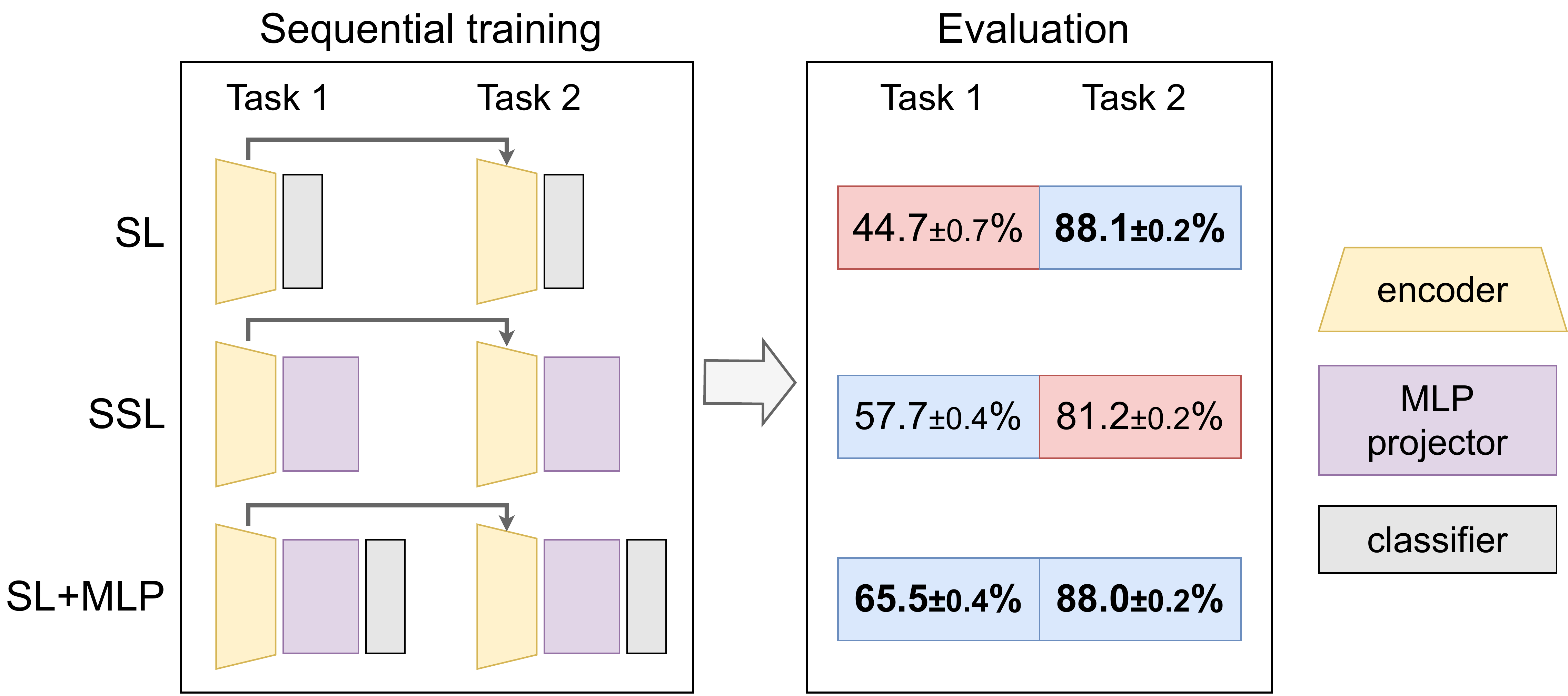}
    \caption{In a two-task continual learning scenario, supervised learning (SL) results in representations that perform well on the second task but poorly on the first task due to high forgetting. On the other hand, representations trained with self-supervised learning (SSL) have higher first-task performance but they underperform on the second task. We show that simple modifications to supervised learning (SL+MLP) yield representations that are superior on the first task and on par with SL on the second task. We report average task-aware k-NN accuracy on 6 different 2-task combinations of CIFAR100, CIFAR10 and SVHN datasets (3 runs for each scenario).}
    \label{fig:teaser}
\end{figure}

In continual learning (CL), the goal of the model is to learn new tasks sequentially. Most of the works focus on supervised continual learning (SCL) for image classification where the learner is provided with labeled training data and the metric of interest is accuracy on all the tasks seen so far.
More recently, unsupervised continual learning (UCL) gained more attention~\cite{fini2022cassle, gomezvilla2022continually, madaan2022lump}. UCL considers the problem of learning robust and general representations on a sequence of tasks, without accessing the data labels. Effective UCL methods would allow the utilization of vast amounts of unlabeled data emerging on a daily basis and continually improve existing models.

A number of recent works study continual learning from a representation learning perspective and show that unsupervised approaches build more robust representations when trained continually~\cite{madaan2022lump, davari2022probing}. More specifically, \cite{madaan2022lump} shows that self-supervised learning (SSL) methods build representations that are more robust to forgetting than supervised learning (SL). \cite{davari2022probing} notice that training SimCLR~\cite{simclr}  have advantageous properties for continual learning compared to supervised training. However, it is still counter-intuitive that access to more information (labels) results in worse representations in continual learning. 

One of the potential reasons is the transferability gap between supervised and unsupervised learning. It was believed that the superior transferability of unsupervised learning can be attributed to a special design of contrastive loss~\cite{zhao2020makes, islam2021broad} or lack of annotations during training~\cite{ericsson2020well, sariyildiz2020concept}. However, recent works~\cite{wang2021revisiting, sariyildiz2023improving} identify that a multi-layer perceptron (MLP) projector commonly used in SSL~\cite{simclr, simsiam, zbontar2021barlow, BYOL} is a crucial component that improves transferability of SSL models. 
Following that finding~\cite{wang2021revisiting, sariyildiz2023improving} use an MLP projector to improve transferability of supervised learning and achieve state-of-the-art transfer learning performance, surpassing unsupervised methods.

In this work, encouraged by these advancements in improving the transferability of supervised models, we revisit supervision for continual representation learning. We argue that additional information (human annotations) should not hurt the quality of representations in continual learning, as suggested by~\cite{madaan2022lump}. Motivated by the latest study on transferability of representations in self-supervised and supervised learning, we aim to improve transferability between tasks in continual learning. We are the first to show that supervised models can continually learn representations of higher quality than self-supervised models when trained with a simple MLP head (see Figure~\ref{fig:teaser}). We identify the crucial role of an MLP projector in representation learning through the perspective of feature transferability, forgetting, and retention for continually trained models. 

The main contributions of this paper are as follows:
\begin{itemize}
    \item We empirically show that SL equipped with a simple MLP projector can learn higher-quality representations than SSL methods in continual finetuning scenarios in both in-distribution and transfer learning scenarios
    \item We show that the use of the MLP projector can be coupled with several continual learning methods, further improving their performance.
    \item We shed light on the reasons behind the strong performance of supervised learning with MLP projector: better transferability, lower forgetting, and increasing diversity of representations.
\end{itemize}

\section{Related Work}

\noindent\textit{\textbf{\textbf{Self-supervised Learning (SSL).}}} 
Learning effective visual representations without annotations is a long-standing problem that aims at leveraging large volumes of unlabeled data.
Recent SSL methods show impressive performance, matching or even exceeding the performance of their supervised equivalents~\cite{simclr,BYOL,zbontar2021barlow, DINO, simsiam}.  The majority of these techniques rely on image augmentation methods to produce multiple views for a given sample. They train a model to be insensitive to these augmentations by ensuring that the network generates similar representations for the views of the same image and different representations for views of other images. In this work, we use BarlowTwins~\cite{zbontar2021barlow} which considers an objective function measuring the cross-correlation matrix between the features and SimCLR~\cite{simclr} which uses contrastive learning based on noise-contrastive estimation. A number of studies~\cite{bordes2023guillotine, simsiam, zbontar2021barlow, jing2021understanding} show that an MLP projector between the encoder and the loss function is a crucial component to prevent the collapse of the representations and improve their transferability. 

\noindent\textit{\textbf{\textbf{Transferable representations.}}} 
\cite{wang2021revisiting} seeks to understand the transferability gap between unsupervised (SSL) and supervised pretraining. They found out that adding a projection network (which is commonly used in SSL) boosts the transferability of the supervised models' features. This was further explored in~\cite{sariyildiz2023improving} and it was shown that it is possible to build representations that are good for both the source and the downstream tasks. In this work, we revisit those findings in the context of models learned on a sequence of tasks. Contrary to the transfer learning literature~\cite{ericsson2020well, sariyildiz2023improving}, which usually focuses on the downstream task performance, we evaluate the model on all tasks during the sequential training. This allows us to gain more insight into learned representations, i.e.~ representation forgetting.

\noindent\textit{\textbf{\textbf{Supervised Continual Learning (SCL).}}} SCL aims to create systems that can acquire the ability to solve novel tasks using new annotated data while retaining the knowledge acquired from previously learned tasks~\cite{parisi2019continual}. A popular formulation of CL is class-incremental learning (CIL)~\cite{masanaLTMBW23, van2019three}
where each task introduces unseen classes that will not occur in the following tasks.
In an exemplar-free setting, the model is not allowed to store any samples from previous tasks which might be important in situations where privacy concerns apply and such a setting remains a great challenge~\cite{exfree}.
A popular strategy is \textit{feature distillation}~\cite{LwF, der}
which minimizes representational changes in subsequent learning stages by enforcing consistent output between the current model and the one trained in the previous task. 
Despite the recent progress in SCL, \cite{stability2023dongwan} highlight the fact that state-of-the-art SCL methods focus on eliminating bias and forgetting on a level of last classification layer. As a result they fail to improve the feature extractor during the continual training which is a main goal of this paper.

\noindent\textit{\textbf{\textbf{Unsupervised Continual Learning (UCL).}}}
Despite the success of SSL methods, they are designed to learn from large static datasets. UCL methods aim to overcome this issue and allow the models to learn from an ever-changing stream of data without excessive memory requirements. 
Recent works~\cite{fini2022cassle, madaan2022lump, gomezvilla2022continually} apply SSL in the UCL setting and claim their superior results for continual representation learning. Most successful methods apply feature distillation through learnable non-linear projector: CaSSLe~\cite{fini2022cassle} distills features outputted by the projector while PFR~\cite{gomezvilla2022continually} distills the features outputted by the backbone.
UCL models are evaluated by measuring their representation strength through linear probing or k-nearest neighbors (k-NN) and this paper follows this evaluation protocol.

\section{Experimental Setup}
\label{sec:exp_setup}

\noindent\textit{\textbf{\textbf{Datasets.}}}
We utilize four different datasets: CIFAR10~\cite{Krizhevsky09learningmultiple} (C10), CIFAR100~\cite{Krizhevsky09learningmultiple} (C100), SVHN~\cite{netzer2011reading} and ImageNet100~\cite{tian2019contrastive} (IN100), 100-class subset of the ILSVRC2012 dataset~\cite{ILSVRC15} with $\approx$ 130k images in high resolution (resized to 224 $\times$ 224). We consider popular settings in continual learning: CIFAR10/5, CIFAR100/5, CIFAR100/20 and ImageNet100/5 sequences, where $D / N$ denotes that dataset $D$ is split into $N$ tasks with an equal number of classes in each task without overlapping ones. To gain further insight, we construct multiple two-task settings where we investigate representation strength and stability. We denote task shift with "$\xrightarrow{}$", e.g. sequence $A\xrightarrow{}B$ means that the model was trained on two tasks, the first one was dataset $A$ and the second one was dataset $B$. We consider C10$\xrightarrow{}$C100 and C100$\xrightarrow{}$C10 scenarios as having low distribution shifts, while C10$\xrightarrow{}$SVHN and SVHN$\xrightarrow{}$C10 scenarios involve higher distribution shifts.
We also perform transfer learning experiments on a set of diverse array of datasets: 
Food101 (Food)~\cite{bossard2014food101}, 
Oxford-IIIT Pets (Pets)~\cite{parkhi2012pets}, 
Oxford Flowers-102 (Flowers)~\cite{nilsback2008flowers}, 
Caltech101 (Caltech)~\cite{li2006caltech}, 
Stanford Cars (Cars)~\cite{krause2013cars}, 
FGVC-Aircraft (Aircrafts)~\cite{maji2013finegrained}, 
Describable Textures (DTD)~\cite{cimpoi14describing} and
Caltech-Birds-200 (Birds)~\cite{wah2011caltech}.

\noindent\textit{\textbf{\textbf{Methods.}} }
We use the following supervised methods:
(1) SL - the standard approach of training a model with linear classification head with a cross-entropy loss function~\cite{masanaLTMBW23}.
(2) SL+MLP - SL with MLP projector added between the backbone and a linear head that is discarded at test-time (see Figure~\ref{fig:teaser}),
(3) t-ReX~\cite{sariyildiz2023improving}, and 
(4) SupCon~\cite{khosla2020supervised}.
Note that SL is the only method that does not utilize an additional MLP projector during training.
For SSL approaches we choose BarlowTwins~\cite{zbontar2021barlow} and SimCLR~\cite{simclr}. Results denoted as SSL were obtained using BarlowTwins. We use ResNet-18~\cite{he2016deep} as a feature extractor network for all the experiments. For CL strategies we use LwF~\cite{LwF}, CaSSLe~\cite{fini2022cassle} and PFR~\cite{gomezvilla2022continually}. Note that we do not include most of the approaches designed for class-incremental learning because most of them fail to improve their feature extractor during continual training~\cite{stability2023dongwan}.

\noindent\textit{\textbf{\textbf{Training.}}} We use the code repository from CaSSLe~\cite{fini2022cassle} and we follow their training procedure. We train SSL models for 500 epochs per task using SGD optimizer with momentum with batch size 256 and cosine learning rate schedule. We adapt the procedure to SL by reducing the number of epochs to 100 per task and the batch size to 64. We tune the learning rate on CIFAR100/5 for each method. We use augmentations from SimCLR~\cite{simclr} for SSL and SupCon and augmentations proposed in~\cite{wang2021revisiting} for SL approaches(SL, SL+MLP and t-ReX). Note that, unless stated otherwise, we investigate continual finetuning scenario and we do not employ any methods for continual learning nor replay buffer.

\noindent\textit{\textbf{\textbf{Evaluation.}}} We use k-NN classifier to evaluate the quality of representations following~\cite{fini2022cassle, madaan2022lump} and Nearest Mean Classifier (NMC) as in~\cite{rebuffi2017_icarl, sdc_2020} to evaluate the stability of representations. We use CKA~\cite{kornblith2019similarity} to measure the similarity between representations of two models. Moreover, we use forgetting ($F$) and forward transfer ($FT$) commonly used in continual learning~\cite{lopez2017gradient}. 
We also measure task exclusion difference $EXC$~\cite{hess2023knowledge} to evaluate the level of retention of task-specific features.
We use subscripts to indicate the evaluation dataset, e.g. $Acc_{C10}$ means "accuracy on C10 dataset". We report means and standard deviations computed across 3 runs unless stated otherwise.

\section{Main Results}

This section presents the experimental results of continual representation learning. 
In Section~\ref{sec:main_results} we present our main results showing that supervised models can outperform self-supervised models in continual representation learning. In Section~\ref{sec:main_transferability} we perform continual transfer learning evaluation that further supports this claim. In Section~\ref{sec:cl_synergy} we combine different models with CL strategies to investigate the synergy between them.
Then we follow with an extensive analysis that sheds light on the reasons for improved performance. Section~\ref{sec:quality_repr} investigates the quality of representations, including forgetting, task exclusion comparison, similarity, and forward transfer. Section~\ref{sec:spectra_repr} presents a spectral analysis of representations.
Finally, in Section~\ref{sec:ablation} we present an ablation study on impact of different components of the projector as well as data efficiency and robustness to label noise.

\newpage

\subsection{Continual representation learning}
\label{sec:main_results}

\begin{wrapfigure}{r}{0.51\textwidth}
    \centering
    \includegraphics[width=0.5\textwidth]{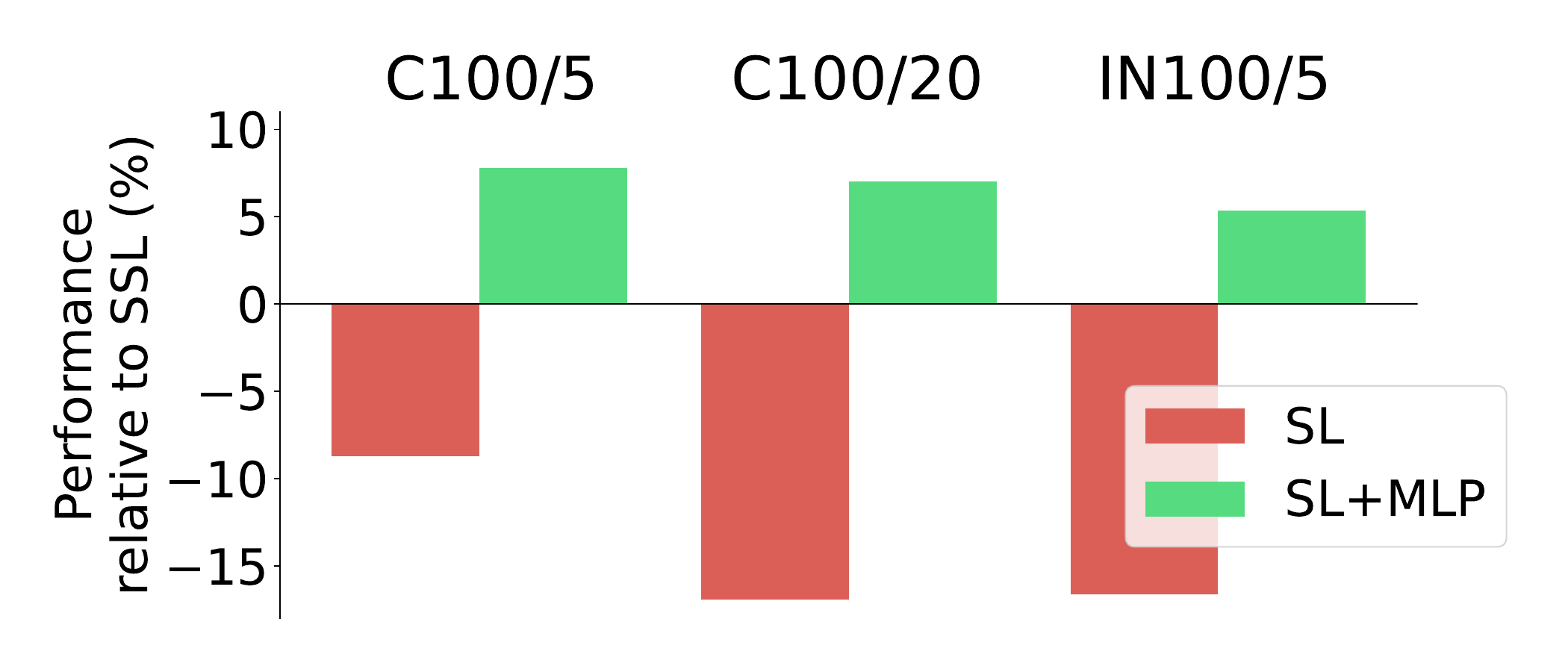}
    \caption{
        SL finetuning underperforms compared to SSL. However, when equipped with the MLP projector it consistently outperforms SSL. We report the difference in k-NN accuracy (\%) between supervised approaches and SSL.
    }
    \label{fig:perf-rel-to-ssl}
\end{wrapfigure}
Figure~\ref{fig:perf-rel-to-ssl} presents our main finding. Namely, we show that supervised models can build stronger representations than self-supervised models under continual finetuning, contrary to previous beliefs~\cite{madaan2022lump}. We identify that the key component to improving the performance of supervised models is an additional MLP projector used during training and discarded afterward - without it, SL significantly underperforms compared to SSL.

\begin{figure}[b]
    \centering
    \includegraphics[width=0.99\textwidth]{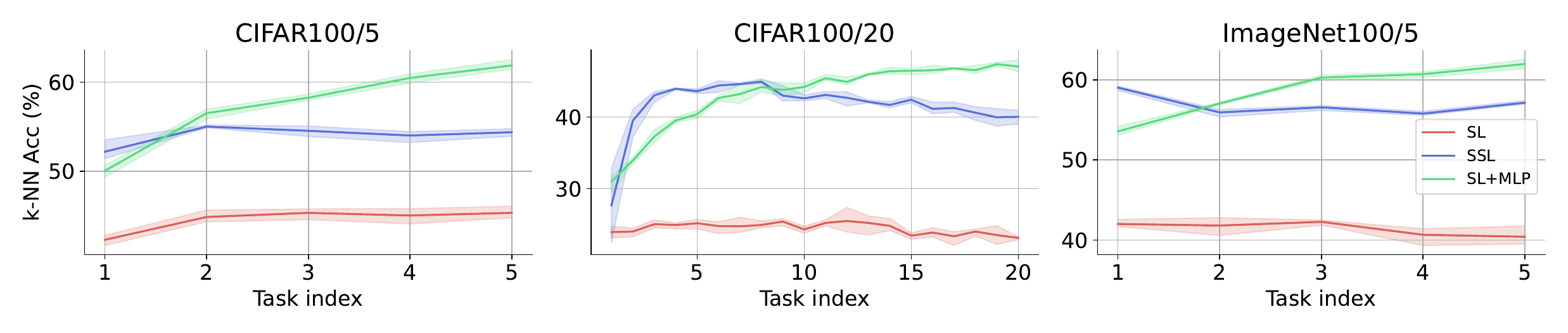}
    \caption{
    SL+MLP: (1) achieves strong performance after the initial task compared to SL which indicates that it produces representations that are transferable to the unseen tasks; (2) is the only method that is able to accumulate knowledge learned on a sequence of tasks. We report task-agnostic k-NN accuracy after each task on the whole dataset (notice that yet unseen tasks are also included in the evaluation).}
    \label{fig:knowledge-accumulation}
\end{figure}

Figure~\ref{fig:knowledge-accumulation} presents the performance of SL, SSL and SL+MLP after each task. We identify two factors contributing to superior results of SL+MLP.
Firstly, we observe that the performance of supervised models after the initial task is largely improved by the addition of the MLP projector, resulting in accuracy close to SSL models. In order to achieve good task-agnostic accuracy on the whole dataset (seen and unseen classes), the model trained on a single task needs to perform well on unseen data. Therefore, we attribute the advantage of SL+MLP to the increased transferability of representations induced by MLP projector, which is in line with~\cite{wang2021revisiting, sariyildiz2023improving}.
Secondly, we notice that SL+MLP is the only method able to incrementally accumulate knowledge and consistently improve performance. This observation is in line with the increasing diversity of features presented in Section~\ref{sec:spectra_repr}.

Table~\ref{tab:longer_sequences} presents extended results including multiple SL and SSL approaches in continual finetuning. We observe that all the supervised methods equipped with the projector significantly outperform simple SL. SL+MLP, t-ReX, and SupCon achieve much higher results than SL in all the finetuning experiments. What is worth noting is the fact that all these methods were trained with different supervised losses: SL+MLP uses cross-entropy, t-ReX uses cosine softmax cross-entropy and SupCon uses supervised contrastive loss. However, they all utilize the MLP projector and all outperform vanilla SL and SSL on most of the datasets.

\begin{table}[t]
  \centering
  \caption{Supervised methods that utilize MLP projector largely outperform vanilla SL. They also outperform SSL on most of the datasets. We report k-NN accuracy of the learned representations. The best result in \textbf{bold} and the second best \underline{underlined}.}
  \scalebox{0.85}{
  \begin{tabular}{lcccc}
    \toprule
      Method &
      \multicolumn{1}{c}{C10/5} &
      \multicolumn{1}{c}{C100/5} &
      \multicolumn{1}{c}{C100/20} &
      \multicolumn{1}{c}{IN100/5} \\
    \midrule
    SL & 59.8$\pm$1.8 & 45.3$\pm$0.7 & 23.1$\pm$0.2 & 40.4$\pm$1.2 \\
    SL+MLP & 65.9$\pm$0.7 & \textbf{61.9$\pm$0.5} & \underline{47.1$\pm$0.7} & \textbf{62.4$\pm$0.4} \\
    t-ReX & 69.3$\pm$1.1 & \underline{59.2$\pm$0.6} & \textbf{50.8$\pm$0.1} & \underline{59.2$\pm$0.6} \\
    SupCon & 60.4$\pm$0.6 & 49.4$\pm$0.3 & 30.0$\pm$0.7 & 57.6$\pm$0.6 \\
    BarlowTwins & \textbf{76.2$\pm$1.2} & 54.1$\pm$0.3 & 40.0$\pm$0.8 & 57.0$\pm$0.4 \\
    SimCLR & \underline{72.4$\pm$1.3} & 48.9$\pm$0.4 & 33.4$\pm$0.5 & 54.7$\pm$0.4 \\
    \bottomrule
  \end{tabular}}
  \label{tab:longer_sequences}
\end{table}

\subsection{Continual transfer learning}
\label{sec:main_transferability}

\begin{figure}[t!]
    \centering
    \includegraphics[width=0.999\textwidth]{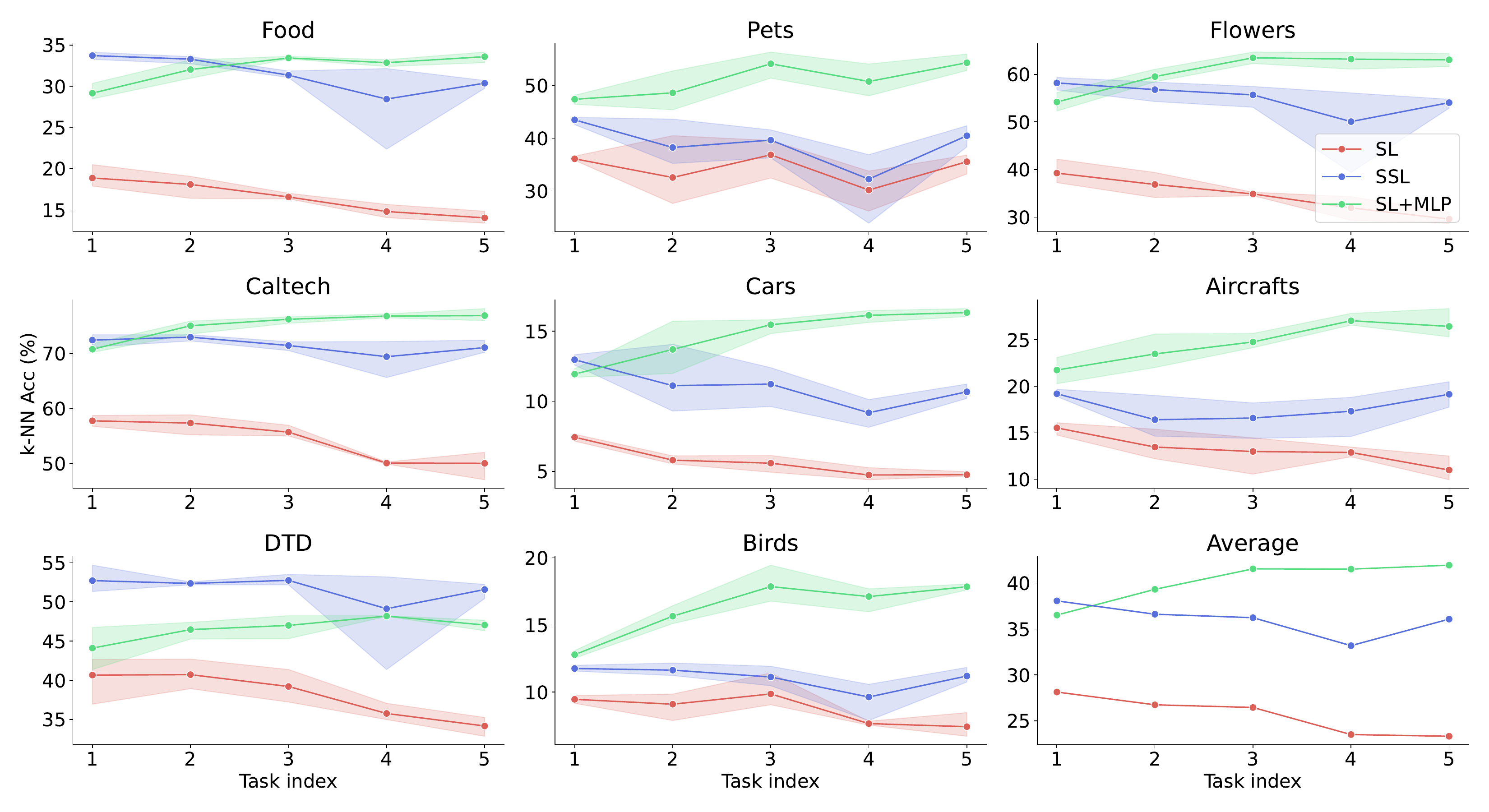}
    \caption{Representations learned by SL+MLP are more transferable than those learned by SL and SSL. They also improve when trained on new tasks. We present the results of the models trained continually on ImageNet/5 and evaluated after each task. We report k-NN accuracy (\%) on a set of 8 diverse downstream classification tasks and an average performance.}
    \label{fig:transfer}
\end{figure}

In this Section we evaluate the continually trained feature extractors on a set of diverse downstream classification tasks described in Section~\ref{sec:exp_setup}: Food, Pets, Flowers, Caltech, Cars, Aircrafts, DTD and Birds. We evaluate the performance on downstream datasets after each task in the continual learning sequence and present the results in Figure~\ref{fig:transfer}. We observe that SL+MLP outperforms the competitors on all but one dataset. Interestingly, the pattern of results is similar to the \textit{in-distribution} results presented in Figure~\ref{fig:knowledge-accumulation}: (1) SL+MLP is on-par with SSL after the first task and significantly outperforms SL, and (2) improves its performance when trained continually. These results highlight the superior transferability of the representations learned by the proposed SL+MLP approach. 

\subsection{Synergy with CL methods}
\label{sec:cl_synergy}

In this Section, we investigate our findings combined with different CL approaches. More precisely, we combine both supervised and self-supervised methods with existing CL methods to verify the synergy between them.

Firstly, we evaluate the resulting combinations \textit{in-distribution} as in Section~\ref{sec:main_results} and present the results in Table~\ref{tab:results_with_cl_methods}. We observe that most of the combinations of methods and CL strategies are outperforming simple finetuning (the only exceptions are some combinations of LwF with supervised methods). Moreover, the conclusions from simple finetuning scenarios still hold -- supervised methods equipped with MLP projectors outperform vanilla SL and SSL. However, with the help of CL methods, the performance gap is much smaller.

\begin{table}[t]
  \centering
  \caption{The effect of MLP projector compounds with the positive effect of CL strategies leading to combinations that outperform SSL on most datasets. 
  We report k-NN accuracy of the learned representations. The best result in each group in \textbf{bold} and the second best \underline{underlined}.
  }
  \scalebox{0.8}{
  \begin{tabular}{llcccccccc}
    \toprule
      Method & CL strategy &
      \multicolumn{2}{c}{C10/5} &
      \multicolumn{2}{c}{C100/5} &
      \multicolumn{2}{c}{C100/20} &
      \multicolumn{2}{c}{IN100/5} \\
    \cmidrule(lr){3-4} \cmidrule(lr){5-6} \cmidrule(lr){7-8} \cmidrule(lr){9-10}
    & & Acc & $\Delta$ & Acc & $\Delta$ & Acc & $\Delta$ & Acc & $\Delta$  \\
    \midrule
    \multicolumn{10}{c}{\textsc{Supervised Continual Learning}} \\
    \midrule
    SL
    & Finetune & 59.8$\pm$1.8 && 45.3$\pm$0.7 && 23.1$\pm$0.2 && 40.4$\pm$1.2 \\
    & LwF & 69.6$\pm$1.1 & \textcolor{Green}{+9.8} & 62.9$\pm$0.1 & \textcolor{Green}{+17.6} & 51.5$\pm$0.2 & \textcolor{Green}{+28.4}& 67.0$\pm$0.1 & \textcolor{Green}{+26.6} \\
    & PFR & 71.0$\pm$2.0 & \textcolor{Green}{+11.2} & \underline{63.3$\pm$0.2} & \textcolor{Green}{+18.0} & 52.3$\pm$1.2 & \textcolor{Green}{+29.2} & 65.0$\pm$0.3 & \textcolor{Green}{+24.6}\\
    \midrule
    SL+MLP
    & Finetune & 65.9$\pm$0.7 && 61.9$\pm$0.5 && 47.1$\pm$0.7 && 62.4$\pm$0.4 \\
    & LwF & 72.6$\pm$3.4 & \textcolor{Green}{+6.7} & 58.7$\pm$0.2 & \textcolor{Red}{-3.2} & 51.9$\pm$0.1 & \textcolor{Green}{+4.8} & 60.4$\pm$0.2 & \textcolor{Red}{-2.0} \\
    & PFR & \underline{76.3$\pm$1.0} & \textcolor{Green}{+10.4} & \textbf{63.6$\pm$0.2} & \textcolor{Green}{+1.7} & \textbf{54.5$\pm$0.2} & \textcolor{Green}{+7.4} & 65.2$\pm$0.1 & \textcolor{Green}{+2.8} \\
    \midrule
    t-ReX
    & Finetune & 69.3$\pm$1.1 && 59.2$\pm$0.6 && 50.8$\pm$0.1 && 59.2$\pm$0.6 \\
    & LwF & 74.5$\pm$0.7 & \textcolor{Green}{+5.2} & 58.3$\pm$0.4 & \textcolor{Red}{-0.9} & 50.4$\pm$0.1 & \textcolor{Red}{-0.4} & 58.6$\pm$1.0 & \textcolor{Red}{-0.6} \\
    & PFR & 75.9$\pm$1.2 & \textcolor{Green}{+6.6} & 60.9$\pm$0.5 & \textcolor{Green}{+1.7} & \underline{53.4$\pm$0.3} & \textcolor{Green}{+2.6} & 63.9$\pm$0.6 & \textcolor{Green}{+4.7} \\
    \midrule 
    SupCon
    & Finetune & 60.4$\pm$0.6 && 49.4$\pm$0.3 && 30.0$\pm$0.7 && 57.6$\pm$0.6 \\
    & CaSSLe & 75.1$\pm$0.4 & \textcolor{Green}{+14.7} & 61.1$\pm$0.2 & \textcolor{Green}{+11.7} & 49.2$\pm$1.2 & \textcolor{Green}{+19.2} & \textbf{70.4$\pm$0.6} & \textcolor{Green}{+12.8} \\
    & PFR & \textbf{78.1$\pm$1.0} & \textcolor{Green}{+17.7} & 57.0$\pm$0.2 & \textcolor{Green}{+7.6} & 51.2$\pm$0.8 & \textcolor{Green}{+21.2} & \underline{68.0$\pm$0.7} & \textcolor{Green}{+10.4} \\
    \midrule
    \multicolumn{10}{c}{\textsc{Unsupervised Continual Learning}} \\
    \midrule
    BarlowTwins
    & Finetune & 76.2$\pm$1.2 && 54.1$\pm$0.3 && 40.0$\pm$0.8 && 57.0$\pm$0.4 \\
    & CaSSLe & \textbf{80.9$\pm$0.2} & \textcolor{Green}{+4.7} & \textbf{58.6$\pm$0.6} & \textcolor{Green}{+4.5} & \underline{49.3$\pm$0.1} & \textcolor{Green}{+9.3} & \textbf{64.9$\pm$0.1} & \textcolor{Green}{+7.9} \\
    & PFR & 78.8$\pm$0.6 & \textcolor{Green}{+2.6} & \underline{57.2$\pm$0.2} & \textcolor{Green}{+3.1} & 46.0$\pm$0.7 & \textcolor{Green}{+6.0} & \underline{61.1$\pm$0.2} & \textcolor{Green}{+4.1} \\
    \midrule
    SimCLR
    & Finetune & 72.4$\pm$1.3 && 48.9$\pm$0.4 && 33.4$\pm$0.5 && 54.7$\pm$0.4 \\
    & CaSSLe & \underline{80.6$\pm$0.5} & \textcolor{Green}{+8.2} & 55.9$\pm$0.5 & \textcolor{Green}{+7.0} & 48.2$\pm$0.4 & \textcolor{Green}{+14.8} & 59.3$\pm$0.5 & \textcolor{Green}{+4.6} \\
    & PFR & 79.2$\pm$0.7 & \textcolor{Green}{+6.8} & 53.8$\pm$0.3 & \textcolor{Green}{+4.9} & \textbf{49.4$\pm$0.1} & \textcolor{Green}{+16.0} & 57.7$\pm$0.2 & \textcolor{Green}{+3.0} \\
    \bottomrule
  \end{tabular}}
  \label{tab:results_with_cl_methods}
\end{table}

We also evaluate transfer learning performance as in Section~\ref{sec:main_transferability} and present the results in Table~\ref{tab:transfer_with_cl}. We observe that CL methods bring an improvement over finetuning in almost all experiments. Similarly to the in-distribution experiments, the best results are obtained by the supervised methods that utilize MLP projector combined with one of CL strategies.

To sum up, models that achieve the best transfer learning performance are those: (1) trained in a supervised way (2) with the use of the MLP projector and (3) coupled with CL strategy based on temporal learnable projection, namely CaSSLe or PFR. These conclusions are coherent with the previous Section where a similar set of models performed the best.

\begin{table}[t]
    \centering
    \caption{
    SL+MLP combined with PFR achieves the best average performance on transfer learning tasks. CL methods also improve the performance of SL and SSL but they achieve worse overall performance.
    Best results in each group (finetuning and continual learning methods) in \textbf{bold}.
    }
    \scalebox{0.79}{
    \begin{tabular}{lcccccccc|c}
    \toprule
    Method & Food & Pets & Flowers & Caltech & Cars & Aircrafts & DTD & Birds & Avg \\
    \midrule
    \multicolumn{10}{c}{\textsc{Finetuning}} \\
    \midrule
    SL & 14.0$\pm$0.6 & 35.6$\pm$1.6 & 29.6$\pm$1.0 & 50.0$\pm$2.1 & 4.8$\pm$0.2 & 11.0$\pm$1.1 & 34.2$\pm$1.0 & 7.4$\pm$0.8 & 23.3 \\
    SSL & 30.4$\pm$0.4 & 40.5$\pm$1.6 & 54.0$\pm$0.8 & 71.1$\pm$1.0 & 10.7$\pm$0.4 & 19.1$\pm$1.1 & \textbf{51.6$\pm$0.8} & 11.2$\pm$0.5 & 36.1 \\
    SL+MLP & \textbf{33.6$\pm$0.5} & \textbf{54.3$\pm$1.3} & \textbf{63.0$\pm$1.1} & \textbf{76.9$\pm$0.9} & \textbf{16.3$\pm$0.2} & \textbf{26.4$\pm$1.4} & 47.1$\pm$0.5 & \textbf{17.8$\pm$0.2} & \textbf{41.9} \\
    \midrule
    \multicolumn{10}{c}{\textsc{Continual learning methods}} \\
    \midrule
    SL+LwF & 30.2$\pm$0.4 & 57.8$\pm$0.8 & 57.9$\pm$0.7 & 72.7$\pm$1.2 & 13.6$\pm$0.3 & 22.2$\pm$0.3 & 46.6$\pm$0.5 & 17.1$\pm$0.4 & 39.8 \\
    SL+PFR & 30.4$\pm$0.6 & 57.4$\pm$0.3 & 58.5$\pm$0.7 & 72.6$\pm$0.2 & 14.0$\pm$0.2 & 21.9$\pm$0.6 & 46.3$\pm$1.1 & 16.3$\pm$0.2 & 39.7 \\
    \midrule
    SSL+PFR & 35.1$\pm$0.3 & 43.4$\pm$1.0 & 59.9$\pm$1.1 & 74.6$\pm$0.4 & 12.2$\pm$0.5 & 19.8$\pm$0.3 & 53.9$\pm$0.7 & 12.4$\pm$0.4 & 38.9 \\
    SSL+CaSSLe & \textbf{39.1$\pm$0.2} & 48.5$\pm$0.7 & 65.1$\pm$0.8 & 75.5$\pm$0.4 & 14.2$\pm$0.6 & 21.1$\pm$0.9 & \textbf{56.4$\pm$0.5} & 13.5$\pm$0.2 & 41.7 \\
    \midrule
    SL+MLP+LwF & 34.7$\pm$0.4 & 55.4$\pm$0.9 & 61.9$\pm$1.3 & 75.3$\pm$0.4 & 15.1$\pm$0.1 & 24.5$\pm$0.5 & 47.3$\pm$1.0 & 16.2$\pm$0.2 & 41.3 \\
    SL+MLP+PFR & 38.0$\pm$0.2 & \textbf{58.8$\pm$0.7} & \textbf{67.8$\pm$0.1} & \textbf{79.3$\pm$0.2} & \textbf{17.5$\pm$0.3} & \textbf{26.2$\pm$0.3} & 49.2$\pm$0.4 & \textbf{18.1$\pm$0.2} & \textbf{44.4} \\
    \bottomrule
    \end{tabular}}
    \label{tab:transfer_with_cl}
\end{table}

\section{Analysis}

\subsection{Quality of representations}
\label{sec:quality_repr}

\begin{table*}[t]
  \centering
\caption{
    We observe high representation forgetting for SL, significantly lower for SSL, and the lowest for SL equipped with MLP projector. SL is able to preserve a small fraction of task-specific features while SL+MLP can retain much more, based on their $EXC$ scores. Surprisingly, SSL achieves negative $EXC$ meaning that pretraining on a given task hurts the performance on this task after the finetuning. We report evaluation on CIFAR10 dataset. The best value between SL, SSL, and SL-MLP in \textbf{bold}.
  }
  \scalebox{0.69}{
  \begin{tabular}{lcccccccccccc}
    \toprule
     Training &
      \multicolumn{4}{c}{SL} &
      \multicolumn{4}{c}{SSL} &
      \multicolumn{4}{c}{SL+MLP}
    \\ \cmidrule(lr){2-5} \cmidrule(lr){6-9} \cmidrule(lr){10-13}
     sequence &
      \multicolumn{1}{c}{$Acc \uparrow$} &
      \multicolumn{1}{c}{$F \downarrow$} &
      \multicolumn{1}{c}{$EXC \uparrow$} &
      \multicolumn{1}{c}{$\text{CKA} \uparrow$} &
      \multicolumn{1}{c}{$Acc \uparrow$} &
      \multicolumn{1}{c}{$F \downarrow$} &
      \multicolumn{1}{c}{$EXC \uparrow$} &
      \multicolumn{1}{c}{$\text{CKA} \uparrow$} &
      \multicolumn{1}{c}{$Acc \uparrow$} &
      \multicolumn{1}{c}{$F \downarrow$} &
      \multicolumn{1}{c}{$EXC \uparrow$} &
      \multicolumn{1}{c}{$\text{CKA} \uparrow$}
\\
    \midrule
    C10 & \textbf{93.7$\pm$0.1} & - & - & - & 88.8$\pm$0.1 & - & - & - & 93.3$\pm$0.1 & - & - & - \\
    \midrule
    C100 & 79.4$\pm$0.3 & - & - & 0.46 & 80.8$\pm$0.1 & - & - & \textbf{0.56} & \textbf{84.5$\pm$0.4} & - & - & 0.49 \\
    C10$\xrightarrow{}$C100 & 81.3$\pm$0.6 & 12.4$\pm$0.6 & 1.9$\pm$0.3 & 0.50 & 79.1$\pm$0.2 & 9.7$\pm$0.3 & -1.8$\pm$0.2 & 0.52 & \textbf{88.8$\pm$0.2} & \textbf{4.5$\pm$0.3} & \textbf{4.3$\pm$0.6} & \textbf{0.57} \\
    \midrule
    SVHN & 27.3$\pm$0.2 & - & - & 0.05 & \textbf{58.6$\pm$1.2} & - & - & \textbf{0.27} & 56.3$\pm$0.2 & - & - & 0.20 \\
    C10$\xrightarrow{}$SVHN & 29.6$\pm$1.1 & 64.1$\pm$1.1 & 2.3$\pm$1.3 & 0.05 & 54.9$\pm$0.7 & 33.8$\pm$0.7 & -3.7$\pm$1.9 & \textbf{0.25} & \textbf{62.7$\pm$0.8} & \textbf{30.6$\pm$0.8} & \textbf{6.4$\pm$1.0} & \textbf{0.25} \\
    \bottomrule  \end{tabular}}
  \label{tab:forgetting}
\end{table*}

\noindent\textit{\textbf{\textbf{Forgetting.}}}
In Table~\ref{tab:forgetting} we observe high representation forgetting for SL, significantly lower for SSL, and the lowest for SL equipped with MLP projector. Low forgetting is an important factor contributing to the superior performance of SL+MLP.

\noindent\textit{\textbf{\textbf{Task exclusion difference.}}}
In the two-task sequence $EXC$ answers the question: \textit{what is the performance gap between the model trained on $B$ and a model trained on a sequence $A \xrightarrow{} B$ when evaluated on $A$}? Results from Table~\ref{tab:forgetting} show that SL achieves small positive $EXC$ meaning that it forgets most features specific to the initial task (but not all of them). SL+MLP achieves the highest $EXC$ which shows that it is able to successfully retain a large portion of task-specific features. Surprisingly, SSL exhibits negative $EXC$. It means that it is better to train SSL model from scratch on another task than to finetune the model pretrained on the task of interest.

\begin{table}[t]
    \centering
    \caption{
    All methods benefit from pretraining on C100 which is semantically close to C10. However, pretraining on semantically distant SVHN hinders the performance of SL. Best method in \textbf{bold}.
    }
    \scalebox{0.85}{
    \begin{tabular}{lccccc}
    \toprule
    \multirow{2}{*}{Method}
     &
      \multicolumn{1}{c}{C10} &
      \multicolumn{2}{c}{C100$\xrightarrow{}$C10} &
      \multicolumn{2}{c}{SVHN$\xrightarrow{}$C10}
      \\ \cmidrule(lr){2-2} \cmidrule(lr){3-4} \cmidrule(lr){5-6} 
     &
      \multicolumn{1}{c}{$Acc_{C10} \uparrow$} &
      \multicolumn{1}{c}{$Acc_{C10} \uparrow$} &
      \multicolumn{1}{c}{$FT_{C10} \uparrow$} &
      \multicolumn{1}{c}{$Acc_{C10} \uparrow$} &
      \multicolumn{1}{c}{$FT_{C10} \uparrow$} \\
    \midrule
    SL & 93.7$\pm$0.1 & 94.3$\pm$0.1 & 0.6$\pm$0.0 & 93.3$\pm$0.1 & -0.4$\pm$0.1 \\
    SSL & 88.8$\pm$0.1 & 89.2$\pm$0.1 & 0.5$\pm$0.2 & 88.5$\pm$0.1 & -0.3$\pm$0.2 \\
    SL+MLP & \textbf{93.3$\pm$0.1} & \textbf{94.3$\pm$0.1} & \textbf{1.0$\pm$0.0} & \textbf{93.2$\pm$0.2} & \textbf{-0.1$\pm$0.1} \\
    \bottomrule
    \end{tabular}}
    \label{tab:forward_transfer}
\end{table}

\noindent\textit{\textbf{\textbf{CKA similarity.}}}
In Table~\ref{tab:forgetting} we report CKA similarity between the models trained on C10 and the rest of the models. We observe that usage of MLP head in SL increases CKA between the C10 model and other models. Moreover, in the case of SL+MLP, the models pretrained on C10 and finetuned on another task have higher similarity to C10 models than the models trained on another dataset from scratch. This is not necessarily the case for SL models. SSL models have the highest CKA scores, however, they usually underperform compared to SL+MLP. This suggests that SSL produces similar features when trained on different datasets but their discriminative power for a classification task is worse than those learned with SL+MLP.

\noindent\textit{\textbf{\textbf{Positive and negative forward transfer.}} }
We present the results of the forward transfer evaluation in Table~\ref{tab:forward_transfer}. All the methods benefit from pretraining on CIFAR100 which is semantically close to CIFAR10. However, pretraining on semantically distant SVHN hinders the performance of SL but it hardly influences the performance of SSL and SL+MLP.

\subsection{Spectra of representations}
\label{sec:spectra_repr}
\begin{figure}[t!]
    \centering
    \includegraphics[width=0.9\textwidth]{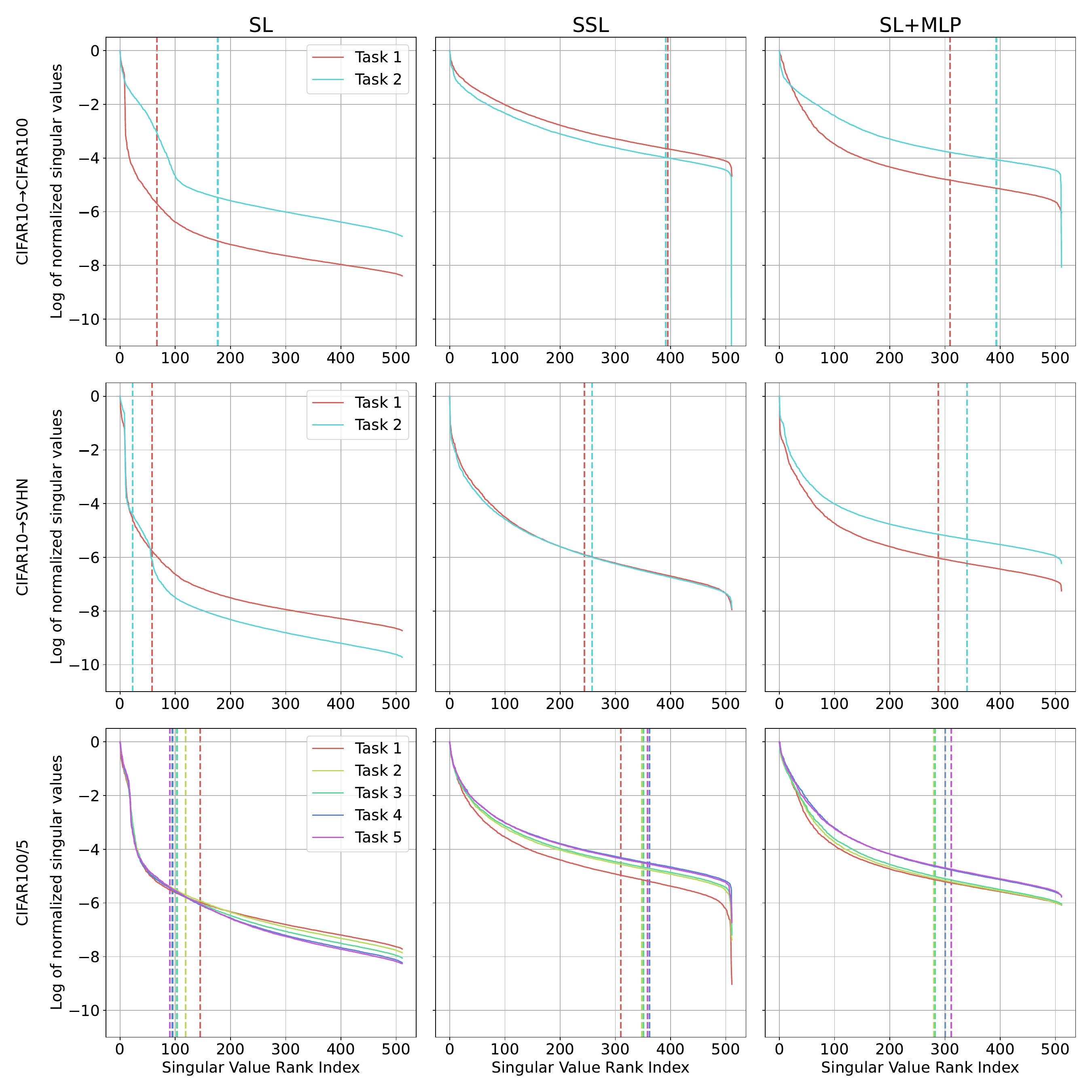}
    \caption{Singular value spectra of 512-dimensional representation space. Representations learned with SL+MLP (right) exhibit desirable properties from the continual learning point of view:
    (1) they consist of a more diverse set of features (contrary to SL, left);
    (2) they improve feature diversity when learning new tasks consistently across all the presented settings. Singular values are ordered descending, are normalized by $\sigma^1$ (the largest singular value) and the scale is logarithmic. Vertical dashed lines denote 95\% of the variance explained. Intuitively, it indicates how many relevant independent features the representation contains.}
    \label{fig:eigenspectra}
\end{figure}

To gain further insight into the properties of continually trained representations, we analyze the spectrum of their covariance matrix. We follow the procedure from~\cite{jing2021understanding}. We gather the representations of the validation set and compute the covariance matrix of the representations, $C$. We perform singular value decomposition of the covariance matrix $C=USV^T$, where $S=diag(\sigma^k)$ and $\sigma^k$ is $k$-th singular value of $C$. We call the representations \textit{diverse} when a large number of principal directions (independent features) is needed to explain most of the variance.
Figure~\ref{fig:eigenspectra} presents how singular value spectra change after each task for different training methods and different sequences of tasks. 

\noindent\textit{\textbf{\textbf{Representation collapse.}}}
Figure~\ref{fig:eigenspectra} reveals that supervised learning exhibits signs of neural collapse~\cite{neural_collapse} - a large fraction of variance is described by a few dimensions roughly equal to the number of classes in the training set. This is an undesirable property in continual representation learning as the representations should be more versatile and useful not only for current but also for past and future tasks. SSL, on the other hand, learns a more diverse set of features resulting in a flatter singular values spectrum. In our experiments adding MLP to SL prevents neural collapse and results in features' properties more similar to SSL.

\noindent\textit{\textbf{\textbf{Evolution of representations.}}}
An important property of representations learned in continual learning is the change in their diversity: the diversity that increases after each task is desired. In Figure~\ref{fig:eigenspectra} we can observe that for SL, the diversity of the features usually decreases, except for C10$\xrightarrow{}$C100 where the increase is caused by a higher number of training classes~\cite{neural_collapse}. For SSL, the diversity increases in the five-task scenario and remains close to constant for two-task settings. SL+MLP is able to improve the diversity of the representations consistently across all the presented scenarios suggesting its superiority in continual representation learning. It may be related to its ability to effectively accumulate knowledge when trained on a sequence of tasks, as presented in Figure~\ref{fig:knowledge-accumulation}.

\subsection{Ablation study}
\label{sec:ablation}

\noindent\textit{\textbf{\textbf{Architecture of the projector.}}}
\begin{table}[t]
  \centering
  \caption{Impact of projector architecture on different methods. Projector architectures are arranged in order of increasing number of parameters. We report k-NN accuracy ($Acc$) and average improvement over the method without projector ($\Delta$). Best results for each method in \textbf{bold}.}
  \scalebox{0.85}{
  \begin{tabular}{llcccc}
    \toprule
      \multirow{1}{*}{Method} & Projector &
      \multicolumn{2}{c}{CIFAR10/5} &
      \multicolumn{2}{c}{CIFAR100/5}
\\ \cmidrule(lr){3-4} \cmidrule(lr){5-6}
    & architecture & $Acc$ & $\Delta$ & $Acc$ & $\Delta$ \\
    \midrule
    SL
    & None & 59.8$\pm$1.8 & & 45.3$\pm$0.7 &\\
    & MLPP~\cite{wang2021revisiting} & 65.9$\pm$0.7 & {\color{Green}+6.1} & \textbf{61.9$\pm$0.5} & \textbf{{\color{Green}+16.6}} \\
    & t-ReX~\cite{sariyildiz2023improving} & \textbf{67.3$\pm$0.1} & \textbf{{\color{Green}+7.5}} & 58.3$\pm$0.2 & {\color{Green} +13.0}\\ 
    \midrule
    t-ReX
    & None & 60.0$\pm$1.2 & & 36.7$\pm$1.1 & \\
    & MLPP~\cite{wang2021revisiting} & 66.7$\pm$1.4 & {\color{Green} +6.7} & 58.1$\pm$0.2 & {\color{Green} +21.4}\\
    & t-ReX~\cite{sariyildiz2023improving} & \textbf{69.3$\pm$1.1} & \textbf{{\color{Green} +9.3}} & \textbf{59.2$\pm$0.6} & \textbf{{\color{Green} +22.5}} \\
    \midrule
    SupCon
    & None & 58.7$\pm$0.4 & & 23.7$\pm$0.5& \\
    & Linear & 61.0$\pm$0.7 & {\color{Green} +2.3} & 37.7$\pm$0.6 & {\color{Green} +14.0}\\
    & SupCon~\cite{khosla2020supervised} & 60.4$\pm$0.6 & {\color{Green} +1.7} & 49.4$\pm$0.3 & {\color{Green} +25.7}\\
    & MLPP~\cite{wang2021revisiting} & \textbf{66.2$\pm$0.7} & \textbf{{\color{Green} +7.5}} & 54.3$\pm$0.5 & {\color{Green} +30.6}\\
    & t-ReX~\cite{sariyildiz2023improving} & 63.5$\pm$0.2 & {\color{Green} +4.8} & \textbf{58.1$\pm$0.1} & \textbf{{\color{Green} +34.4}} \\
    \bottomrule
  \end{tabular}}
  \label{tab:proj-types}
\end{table}
In Table~\ref{tab:proj-types} we examine the impact of different projector architectures on supervised learning methods. All the methods achieve the worst performance when not using a projector. Moreover, all the methods achieve similarly high accuracy when with a suitable projector. These results highlight the importance of MLP projector for supervised continual representation learning and diminish the importance of other factors (\textit{e.g.} loss function) on the final performance.

To gain a deeper insight into the impact of different components of the projector we systematically investigate the impact of the depth and width of the projector. We follow the MLPP~\cite{wang2021revisiting} architecture for the projector, using a linear layer, batch normalization (BN), and ReLU activation, which we call a \textit{block} with a hidden dimension $d_h$. This block is followed by an output linear layer with a hidden dimension $d_o$, and then a classification layer. The projector contains $n$ blocks, where $n=0$ means a linear projector. We present the results in Table~\ref{tab:proj-depth-width}. We observe that the width and depth of the projector have a very limited impact on the final performance as long as the projector has at least one block. Therefore, we further decompose the projector starting from an MLP with a single block ($d_h=4096$, $d_o=512$) and remove basic components. We present the results in Table~\ref{tab:proj-components}. We observe that BN has the highest positive impact on performance. However, using BN without the linear layer and ReLU lags 4 p.p. behind the performance of the full block, highlighting the importance of all the components.

\begin{figure}[t]
  \begin{minipage}{.62\linewidth}
    \centering
    \caption{One MLP block brings significant improvements and the next blocks have minor effect. Impact of projector depth and width on CIFAR100/5. Note that $d_h$ does not apply to linear projectors.}
    \scalebox{0.85}{
    \begin{tabular}{l|cc|cc|cc|cc}
         \toprule
         $n=$ & \multicolumn{2}{c|}{0 (linear)} & \multicolumn{2}{c|}{1} & \multicolumn{2}{c|}{2} & \multicolumn{2}{c}{3} \\
         $h_o=$ & 512 & 2048 & 512 & 2048 & 512 & 2048 & 512 & 2048 \\
         \midrule
         $d_h=512$ & \multirow{4}{*}{\shortstack[c]{46.74}} & \multirow{4}{*}{47.45} & 60.06 & 60.88 & 60.61 & 60.45 & 60.59 & 60.93 \\
         $d_h=1024$ &&& 60.61 & 60.50 & 61.04 & 60.61 & 60.84 & 61.25 \\
         $d_h=2048$ &&& 60.54 & 60.88 & 61.42 & 61.05 & 60.97 & 61.00 \\
         $d_h=4096$ &&& 61.46 & 61.10 & 60.95 & 61.46 & 61.09 & 60.91 \\
         \bottomrule
    \end{tabular}}
    \label{tab:proj-depth-width}
  \end{minipage}
  \hfill
  \begin{minipage}{.36\linewidth}
    \centering
    \caption{BatchNorm seems to have the highest positive impact on the final performance. Impact of block components on CIFAR100/5.}
    \scalebox{0.85}{
    \begin{tabular}{ll|cccc}
         \toprule
         \multicolumn{2}{l|}{Input Lin.} & \xmark & \xmark & \cmark & \cmark \\
         \multicolumn{2}{l|}{ReLU} & \xmark & \cmark & \xmark & \cmark \\
         \midrule
         \multirow{2}{*}{BN} & \xmark & 46.74 & 49.06 & 47.59 & 53.87 \\
         & \cmark & 57.33 & 59.15 & 58.47 & 61.46 \\
         \bottomrule
    \end{tabular}}
    \label{tab:proj-components}
  \end{minipage}
\end{figure}

\begin{figure}[t]
    \centering
    \includegraphics[width=0.49\textwidth]{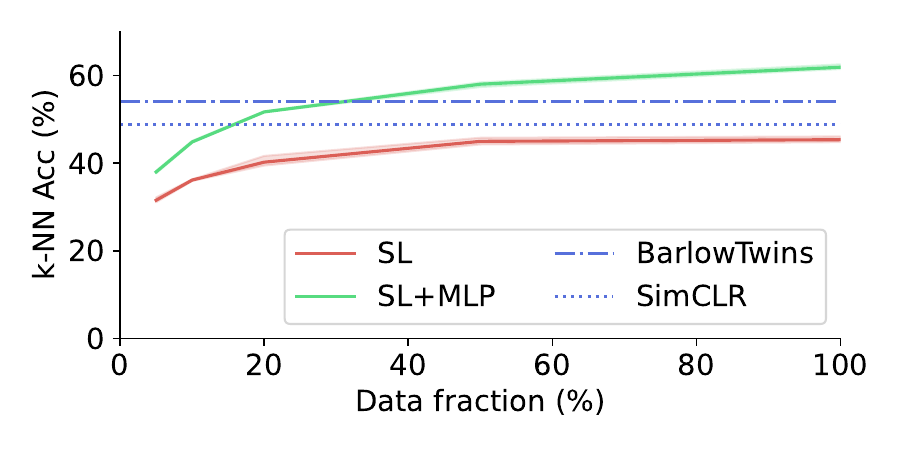}
    \includegraphics[width=0.49\textwidth]{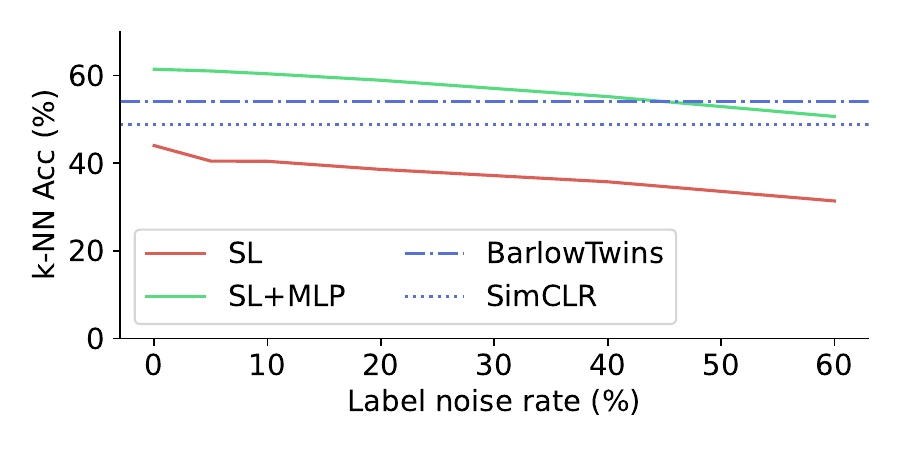}
    \caption{SL+MLP outperforms SSL and SL approaches: (left) trained on full dataset even with only 30\% of data available and (right) with 40\% label noise. We present the accuracy (\%) after the final task on CIFAR100/5 for different data fractions (left) and varying label noise level (right).} 
    \label{fig:data-percentage-and-label-noise}
\end{figure}

\noindent\textit{\textbf{\textbf{Data efficiency.}}}
SL with MLP projector outperforms SSL in continual representation learning when having access to the same amount of data. However, in real-world scenarios, SSL is able to utilize vast amounts of unlabeled data while SL needs costly data annotations. Therefore, we examine how SL behaves when trained on a fraction of data available for SSL in each task to simulate limited access to labeled data. We restrict the data and select a fraction of it (equal for each task).
We present the results in Figure~\ref{fig:data-percentage-and-label-noise} (left). SL+MLP surpasses SimCLR with less than 20\% of data and BarlowTwins with about 30\% of data.

\noindent
\textbf{\textit{Robustness to mislabelling.}}
We evaluate the robustness of our claims against noisy labels following previous study~\cite{Kim_2021_ICCV} and randomly change the labels of the fraction of samples in the train dataset. Figure~\ref{fig:data-percentage-and-label-noise} (right) presents the results on CIFAR100/5. We observe that SL+MLP outperforms SSL even with 40\% label noise. It validates the efficacy of SL+MLP in the presence of imperfect annotations.

\section{Discussion and limitations}

Although supervised learning with the MLP projection head seems to be more effective in continual representation learning, it comes at a price. SL requires mundane image labeling of the whole dataset which can be costly and impractical at scale. Self-supervised learning, on the other hand, is not dependent on image annotations and, therefore, can operate on a vast amount of unlabeled data.

However, SSL faces its own limitations. Firstly, most SSL approaches depend on strong image augmentations and learn representations that are invariant to them~\cite{simclr, zbontar2021barlow, DINO}. This can hinder the performance on the downstream tasks which require attention to the traits that it has been trained to be invariant to~\cite{xiao2021contrastive}. Moreover, SSL usually requires longer training which increases computational requirements in comparison to SL.
It is also worth noting that both SL+MLP and SSL introduce additional costs to the model during the training, as both introduce MLP projector that requires more computational requirements. However, at test time every method operates at the same number of parameters, as we discard MLP projectors after training.

Furthermore, it's worth re-emphasizing that this work focuses on continual representation learning. While we utilize data from previous tasks to construct k-NN and nearest mean classifiers for evaluating learned representations, our primary objective is not centered on the continual approach to the downstream task (classification). We are not delving into class-incremental learning, a prevalent continual learning setting. Nonetheless, our analysis of representation strength and stability can offer valuable insights into continual learning dynamics, potentially aiding in the creation of more effective algorithms for continual downstream task solutions.

\section{Conclusions}
In this work, we are first to show that supervised learning can significantly outperform self-supervised learning in continual representation learning. We achieve it by equipping SL with a simple MLP projector discarded after the training, following the common practice from SSL. We show that SL+MLP can be successfully coupled with several continual learning strategies, further improving the performance. Finally, we shed some light on the reasons for improved performance when using MLP with SL: better transferability, lower forgetting, and higher diversity of learned features.

\section*{Acknowledgments}
Daniel Marczak is supported by National Centre of Science (NCN, Poland) Grant No. 2021/43/O/ST6/02482.
This research was partially funded by National Science Centre, Poland, grant no 2020/39/B/ST6/01511, grant no 2022/45/B/ST6/02817, and grant no 2023/51/D/ST6/02846.
Bartłomiej Twardowski acknowledges the grant RYC2021-032765-I.
This paper has been supported by the Horizon Europe Programme (HORIZON-CL4-2022-HUMAN-02) under the project "ELIAS: European Lighthouse of AI for Sustainability", GA no. 101120237.
We gratefully acknowledge Polish high-performance computing infrastructure PLGrid (HPC Center: ACK Cyfronet AGH) for providing computer facilities and support within computational grant no. PLG/2023/016393.

\bibliographystyle{splncs04}
\bibliography{main}

\begin{thebibliography}{10}
\providecommand{\url}[1]{\texttt{#1}}
\providecommand{\urlprefix}{URL }
\providecommand{\doi}[1]{https://doi.org/#1}

\bibitem{bordes2023guillotine}
Bordes, F., Balestriero, R., Garrido, Q., Bardes, A., Vincent, P.: Guillotine
  regularization: Why removing layers is needed to improve generalization in
  self-supervised learning. TMLR  (2023)

\bibitem{bossard2014food101}
Bossard, L., Guillaumin, M., Van~Gool, L.: Food-101 -- mining discriminative
  components with random forests. In: ECCV (2014)

\bibitem{DINO}
Caron, M., Touvron, H., Misra, I., Jegou, H., Mairal, J., Bojanowski, P.,
  Joulin, A.: Emerging properties in self-supervised vision transformers. ICCV
  (2021)

\bibitem{simclr}
Chen, T., Kornblith, S., Norouzi, M., Hinton, G.E.: A simple framework for
  contrastive learning of visual representations. In: ICML (2020)

\bibitem{simsiam}
Chen, X., He, K.: Exploring simple siamese representation learning. CVPR
  (2020)

\bibitem{cimpoi14describing}
Cimpoi, M., Maji, S., Kokkinos, I., Mohamed, S., , Vedaldi, A.: Describing
  textures in the wild. In: CVPR (2014)

\bibitem{davari2022probing}
Davari, M.R., Asadi, N., Mudur, S., Aljundi, R., Belilovsky, E.: Probing
  representation forgetting in supervised and unsupervised continual learning.
  In: CVPR (2022)

\bibitem{ericsson2020well}
Ericsson, L., Gouk, H., Hospedales, T.M.: How well do self-supervised models
  transfer? CVPR  (2020)

\bibitem{li2006caltech}
Fei-Fei, L., Fergus, R., Perona, P.: One-shot learning of object categories.
  IEEE TPAMI  (2006)

\bibitem{fini2022cassle}
Fini, E., da~Costa, V.G.T., Alameda{-}Pineda, X., Ricci, E., Alahari, K.,
  Mairal, J.: Self-supervised models are continual learners. In: CVPR (2022)

\bibitem{gomezvilla2022continually}
Gomez-Villa, A., Twardowski, B., Yu, L., Bagdanov, A.D., van~de Weijer, J.:
  Continually learning self-supervised representations with projected
  functional regularization. CVPR Workshops  (2021)

\bibitem{BYOL}
Grill, J.B., Strub, F., Altch'e, F., Tallec, C., Richemond, P.H., Buchatskaya,
  E., Doersch, C., Pires, B.., Guo, Z., Azar, M.G., Piot, B., Kavukcuoglu, K.,
  Munos, R., Valko, M.: Bootstrap your own latent: A new approach to
  self-supervised learning. NeurIPS  (2020)

\bibitem{he2016deep}
He, K., Zhang, X., Ren, S., Sun, J.: Deep residual learning for image
  recognition. In: CVPR (2016)

\bibitem{hess2023knowledge}
Hess, T., Verwimp, E., van~de Ven, G.M., Tuytelaars, T.: Knowledge accumulation
  in continually learned representations and the issue of feature forgetting.
  TMLR  (2024)

\bibitem{islam2021broad}
Islam, A., Chen, C.F., Panda, R., Karlinsky, L., Radke, R., Feris, R.: A broad
  study on the transferability of visual representations with contrastive
  learning. ICCV  (2021)

\bibitem{jing2021understanding}
Jing, L., Vincent, P., LeCun, Y., Tian, Y.: Understanding dimensional collapse
  in contrastive self-supervised learning. In: ICLR (2022)

\bibitem{khosla2020supervised}
Khosla, P., Teterwak, P., Wang, C., Sarna, A., Tian, Y., Isola, P., Maschinot,
  A., Liu, C., Krishnan, D.: Supervised contrastive learning. NeurIPS  (2020)

\bibitem{Kim_2021_ICCV}
Kim, C.D., Jeong, J., Moon, S., Kim, G.: Continual learning on noisy data
  streams via self-purified replay. In: ICCV (2021)

\bibitem{stability2023dongwan}
Kim, D., Han, B.: On the stability-plasticity dilemma of class-incremental
  learning. CVPR  (2023)

\bibitem{kornblith2019similarity}
Kornblith, S., Norouzi, M., Lee, H., Hinton, G.E.: Similarity of neural network
  representations revisited. ICML  (2019)

\bibitem{krause2013cars}
Krause, J., Stark, M., Deng, J., Fei-Fei, L.: 3d object representations for
  fine-grained categorization. In: 4th International IEEE Workshop on 3D
  Representation and Recognition (3dRR-13) (2013)

\bibitem{Krizhevsky09learningmultiple}
Krizhevsky, A.: Learning multiple layers of features from tiny images.
  University of Toronto  (2009)

\bibitem{LwF}
Li, Z., Hoiem, D.: Learning without forgetting. IEEE TPAMI  (2018)

\bibitem{lopez2017gradient}
Lopez-Paz, D., Ranzato, M.: Gradient episodic memory for continual learning.
  NeurIPS  (2017)

\bibitem{madaan2022lump}
Madaan, D., Yoon, J., Li, Y., Liu, Y., Hwang, S.J.: Representational continuity
  for unsupervised continual learning. In: ICLR (2022)

\bibitem{maji2013finegrained}
Maji, S., Rahtu, E., Kannala, J., Blaschko, M., Vedaldi, A.: Fine-grained
  visual classification of aircraft (2013)

\bibitem{masanaLTMBW23}
Masana, M., Liu, X., Twardowski, B., Menta, M., Bagdanov, A.D., van~de Weijer,
  J.: Class-incremental learning: Survey and performance evaluation on image
  classification. IEEE TPAMI  (2023)

\bibitem{netzer2011reading}
Netzer, Y., Wang, T., Coates, A., Bissacco, A., Wu, B., Ng, A.Y.: Reading
  digits in natural images with unsupervised feature learning  (2011)

\bibitem{nilsback2008flowers}
Nilsback, M.E., Zisserman, A.: Automated flower classification over a large
  number of classes. In: Indian Conference on Computer Vision, Graphics and
  Image Processing (2008)

\bibitem{neural_collapse}
Papyan, V., Han, X.Y., Donoho, D.L.: Prevalence of neural collapse during the
  terminal phase of deep learning training. PNAS  (2020)

\bibitem{parisi2019continual}
Parisi, G.I., Kemker, R., Part, J.L., Kanan, C., Wermter, S.: Continual
  lifelong learning with neural networks: A review. {Neural Networks}  (2019)

\bibitem{parkhi2012pets}
Parkhi, O.M., Vedaldi, A., Zisserman, A., Jawahar, C.V.: Cats and dogs. In:
  CVPR (2012)

\bibitem{rebuffi2017_icarl}
Rebuffi, S., Kolesnikov, A., Sperl, G., Lampert, C.H.: icarl: Incremental
  classifier and representation learning. In: CVPR (2017)

\bibitem{ILSVRC15}
Russakovsky, O., Deng, J., Su, H., Krause, J., Satheesh, S., Ma, S., Huang, Z.,
  Karpathy, A., Khosla, A., Bernstein, M., Berg, A.C., Fei-Fei, L.: {ImageNet
  Large Scale Visual Recognition Challenge}. IJCV  (2015)

\bibitem{sariyildiz2023improving}
Sariyildiz, M.B., Kalantidis, Y., Alahari, K., Larlus, D.: No reason for no
  supervision: Improved generalization in supervised models. In: ICLR (2023)

\bibitem{sariyildiz2020concept}
Sariyildiz, M.B., Kalantidis, Y., Larlus, D., Karteek, A.: Concept
  generalization in visual representation learning. ICCV  (2021)

\bibitem{exfree}
Smith, J.S., Tian, J., Halbe, S., Hsu, Y., Kira, Z.: A closer look at
  rehearsal-free continual learning. In: CVPR Workshops (2023)

\bibitem{tian2019contrastive}
Tian, Y., Krishnan, D., Isola, P.: Contrastive multiview coding. ECCV  (2019)

\bibitem{van2019three}
van~de Ven, G.M., Tuytelaars, T., Tolias, A.S.: Three types of incremental
  learning. Nature Machine Intelligence  (2022)

\bibitem{wah2011caltech}
Wah, C., Branson, S., Welinder, P., Perona, P., Belongie, S.: The caltech-ucsd
  birds-200-2011 dataset (2011)

\bibitem{wang2021revisiting}
Wang, Y., Tang, S., Zhu, F., Bai, L., Zhao, R., Qi, D., Ouyang, W.: Revisiting
  the transferability of supervised pretraining: an mlp perspective. CVPR
  (2021)

\bibitem{xiao2021contrastive}
Xiao, T., Wang, X., Efros, A.A., Darrell, T.: What should not be contrastive in
  contrastive learning. In: ICLR (2021)

\bibitem{der}
Yan, S., Xie, J., He, X.: {DER:} dynamically expandable representation for
  class incremental learning. In: CVPR (2021)

\bibitem{sdc_2020}
Yu, L., Twardowski, B., Liu, X., Herranz, L., Wang, K., Cheng, Y., Jui, S.,
  van~de Weijer, J.: Semantic drift compensation for class-incremental
  learning. In: CVPR (2020)

\bibitem{zbontar2021barlow}
Zbontar, J., Jing, L., Misra, I., LeCun, Y., Deny, S.: {Barlow Twins:
  Self-Supervised Learning via Redundancy Reduction}. ICML  (2021)

\bibitem{zhao2020makes}
Zhao, N., Wu, Z., Lau, R.W.H., Lin, S.: What makes instance discrimination good
  for transfer learning? ICLR  (2020)

\end{thebibliography}

\appendix
\clearpage
\setcounter{page}{1}

\section{Implementation details}
\label{sec:apdx_impl_details}

\subsection{CL strategies}

\textbf{LwF}~\cite{LwF} is a classic SCL method for feature distillation. It distills the logits of the frozen network trained on previous tasks using cross-entropy loss. We pair it with SL methods that train with cross-entropy loss. We use the implementation from~\cite{masanaLTMBW23}.

\noindent
\textbf{CaSSLe}~\cite{fini2022cassle} is a method for self-supervised continual learning that utilizes a learnable MLP to project past features onto the new latent space for improved feature distillation. The distillation is performed on the outputs from the SSL projector with the loss function of a particular SSL method. Because of reliance on SSL-specific components, namely the projector and loss function, we do not pair CaSSLe with supervised approaches, except for SupCon which loss and architecture closely resemble SSL. We follow an official implementation of CaSSLe.

\noindent
\textbf{PFR}~\cite{gomezvilla2022continually} realizes a similar idea to CaSSLe. It also uses a learnable MLP projector to enhance feature distillation. However, it uses cosine similarity as a loss function and performs distillation on the outputs of the backbone network. Therefore, we pair it with both SL and SSL approaches as it does not rely on SSL-specific components.
We present the chosen values of regularization hyperparameter $\lambda$  in Table~\ref{tab:pfr_lambda}. We selected the best $\lambda \in \{1, 3, 10, 15, 25\}$ separately for each method and dataset.

\begin{table}[!htbp]
  \centering
  \caption{PFR regularization hyperparameter $\lambda$ for different methods and datasets.}
  \scalebox{0.9}{
  \begin{tabular}{lcccc}
    \toprule
      \multirow{1}{*}{Method} &
      \multicolumn{1}{c}{C10/5} &
      \multicolumn{1}{c}{C100/5} &
      \multicolumn{1}{c}{C100/20} &
      \multicolumn{1}{c}{IN100/5}
\\
    \midrule
    SL & 1.0 & 10.0 & 10.0 & 15.0 \\
    SL+MLP & 3.0 & 3.0 & 10.0 & 1.0 \\
    t-ReX & 3.0 & 3.0 & 10.0 & 1.0 \\
    SupCon & 3.0 & 10.0 & 25.0 & 10.0 \\
    BarlowTwins & 25.0 & 25.0 & 25.0 & 25.0 \\
    SimCLR & 3.0 & 3.0 & 15.0 & 3.0 \\
    \bottomrule
  \end{tabular}}
  \label{tab:pfr_lambda}
\end{table}

\subsection{k-NN evaluation}
Each model is evaluated with a k-nearest neighbour classifier after training each task (offline evaluation). Moreover, we perform some experiments where we use k-nn evaluation after each epoch (online evaluation for Figure~\ref{fig:feature-retention-extended}).

For online evaluation, we perform extensive hyperparameter search and report results obtained by the best probe. We explore the following hyperparameters:
\begin{itemize}
    \item $k \in \{5, 10, 20, 50, 100, 200\}$ - number of considered neighbours;
    \item distance function - we consider either euclidean distance or cosine similarity;
    \item temperature $T \in \{0.02, 0.05, 0.07, 0.1, 0.2, 0.5 \}$ used only with cosine distance;
\end{itemize}
resulting in 42 k-NN probes per one offline evaluation.

For online evaluation, we use a fixed hyperparameter set: $k=20$, cosine distance, and $T=0.07$. This k-NN configuration often turns out to be one of the best in offline evaluation.

\subsection{Architectures of the projector}
\label{sec:apdx_proj_arch}

\begin{figure*}[h!]
    \centering
    \includegraphics[width=0.98\textwidth]{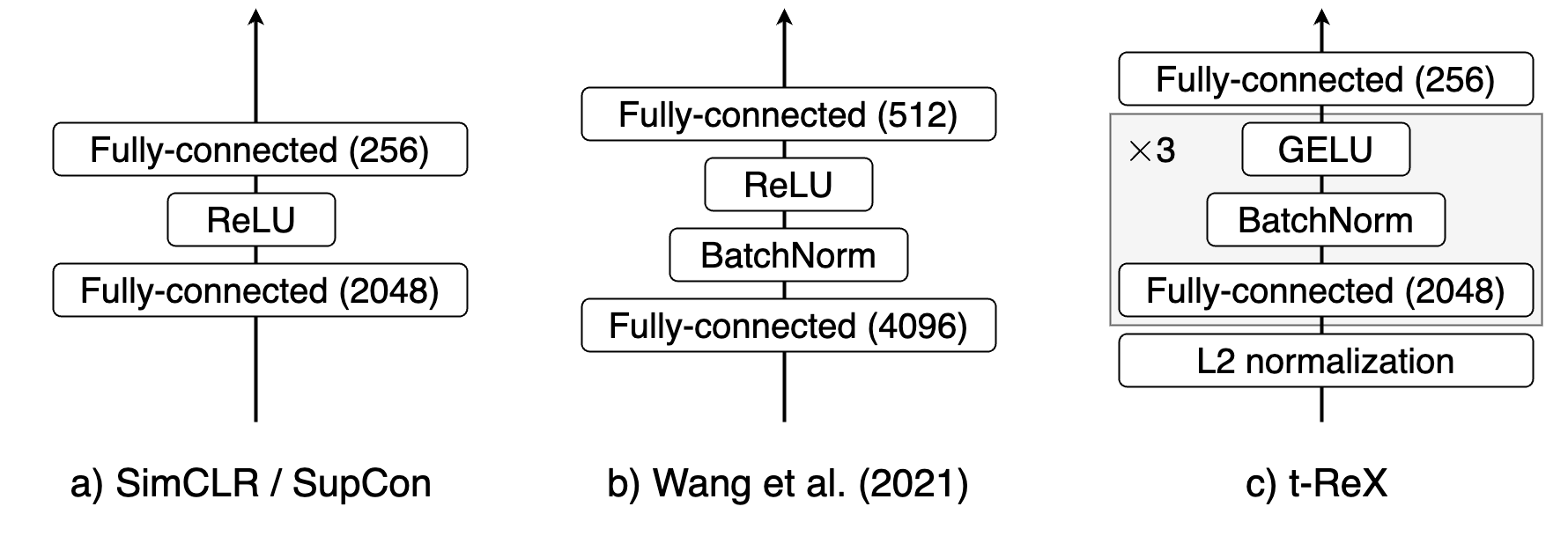}
    \caption{Architectures of the projectors used by different methods.} 
    \label{fig:projector-archs}
\end{figure*}

In Figure~\ref{fig:projector-archs} we present the architectures of the projectors proposed by SimCLR~\cite{simclr}, t-ReX~\cite{sariyildiz2023improving} and by~\cite{wang2021revisiting}. The results of different methods paired with these projectors are presented in Table~\ref{tab:proj-types} in the main paper.

\section{Extended analysis}

\subsection{Initialization of MLP projector}
Is it better to randomly reinitialize the projector after each task or is it better to start from the weights of the projector learned on the previous tasks? The results reported in Table~\ref{tab:mlp-init} suggest that for SL it is better to randomly reinitialize the MLP projector. SSL methods, however, tend to perform slightly better when the projector for a new task is initialized with the previously learned projector. We suspect that in SL the projector encompasses task-specific knowledge which interferes with learning new tasks. On the other hand, in SSL the MLP is responsible for projecting the representations into the space where an invariance to augmentations is enforced which is less task-specific than classification.

\begin{table}[h!]
  \captionof{table}{
  SL benefits from resetting the MLP projector and SSL methods tend to perform slightly better when starting from the weights of the projector learned on previous tasks. We report k-NN accuracy (\%) after the final task. Better initialization method in \textbf{bold}.
  }
  \centering
  \scalebox{0.9}{
  \begin{tabular}{llcc}
    \toprule
      \multirow{1}{*}{Method} & MLP init &
      \multicolumn{1}{c}{CIFAR10/5} &
      \multicolumn{1}{c}{CIFAR100/5} \\
    \midrule
    SL+MLP
    & Reset & \textbf{65.9$\pm$0.7} & \textbf{61.9$\pm$0.5} \\
    & Previous & 65.0$\pm$1.5 & 60.1$\pm$0.2 \\
    \midrule
    BarlowTwins
    & Reset & 76.1$\pm$0.5 & \textbf{54.9$\pm$0.3} \\
    & Previous & \textbf{76.2$\pm$1.2} & 54.1$\pm$0.3 \\
    \midrule
    SimCLR
    & Reset & 72.0$\pm$1.6 & 47.4$\pm$0.2 \\
    & Previous & \textbf{72.4$\pm$1.3} & \textbf{48.9$\pm$0.4} \\
    \bottomrule
  \end{tabular}}
  \label{tab:mlp-init}
\end{table}

\subsection{Stability of representations}
\label{sec:stability_repr}

We define representations as \textit{stable} when they do not drift in the representation space when the network is trained on a new task.
The stability of representations is a desired property of SCL models as stable representations facilitate continual training of a classifier~\cite{sdc_2020}. On the other hand, UCL evaluation only measures the representations' strength and the relationship of stability and strength of representations is not obvious. One can imagine both stable and unstable representations can improve strength during continual training.

In this section, we evaluate the stability of representations of SL and SSL models. We use nearest mean classifier (NMC) accuracy to measure it in the context of SCL. After the first task, we calculate prototypes of each class as a mean feature of all the samples of this class. We evaluate the model and save the prototypes. Then, we train on the second task and evaluate the model using saved prototypes. We use the accuracy obtained by classification using old prototypes as a proxy of the stability of the representations. In the case of perfectly stable representations, both evaluations would result in the same accuracy while perfectly unstable representations would cause accuracy to drop to a random guess level. Moreover, we evaluate the updated model using prototypes recalculated on previous data (not allowed in continual learning) to provide an upper bound.

\begin{figure}[h!]
    \centering
    \includegraphics[width=0.99\textwidth]{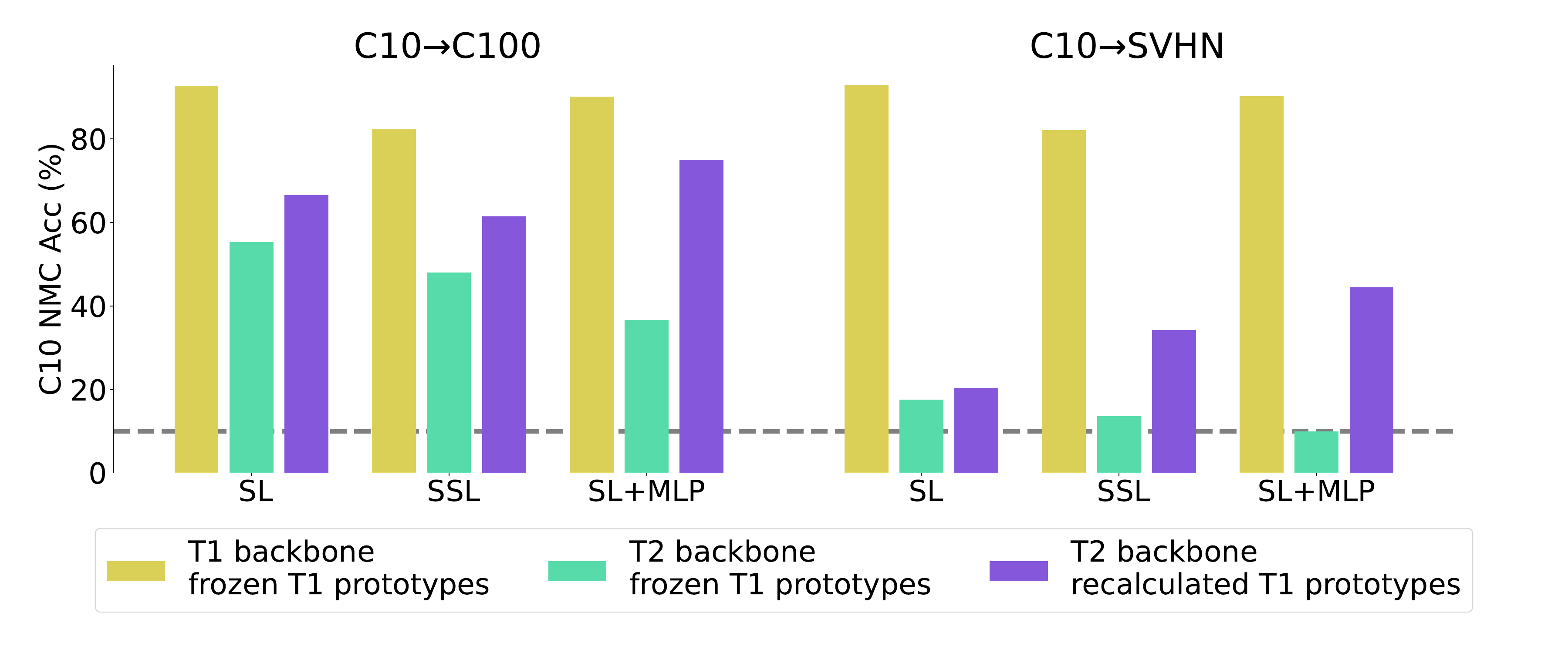}
    \caption{Task aware NMC accuracy on CIFAR10 dataset for supervised and self-supervised models trained on different sequences of tasks. After training on CIFAR10 (T1), both SL and SSL models achieve high NMC performance (yellow). After training the second task (T2), the nearest mean classification using old prototypes results in performance degradation (green). We calculate \emph{an upper-bound} accuracy after training on the second task by recalculating the prototypes using old data and a new backbone (purple). Note that it is not possible in the CL scenario as old data is inaccessible. Gray dotted line marks random guess performance.} 
    \label{fig:repr_drift}
\end{figure}

The results are presented in Figure~\ref{fig:repr_drift}. Representations of all the methods are not stable in high distribution shift scenario C10$\xrightarrow{}$SVHN. They achieve random guess accuracy when utilizing saved (old) prototypes.
However, in a low distribution shift scenario, C10$\xrightarrow{}$C100, SL achieves 55.3\% accuracy using old prototypes (11.3\% below upper bound performance) while
SSL achieves 48.0\% (14.5\% below upper-bound) and SL+MLP achieves only 36.7\% (38.3\% below upper-bound). Note that performance degradation can be only partially attributed to forgetting of representations as the upper-bound performance is still high after training on the second task for most of the methods. These results suggest that there exists a trade-off between the stability of representations and the expressiveness of representations trained continually as methods that build stronger representations tend to have lower stability.

\subsection{Impact of training length}
\begin{figure}[h!]
    \centering
    \includegraphics[width=0.999\textwidth]{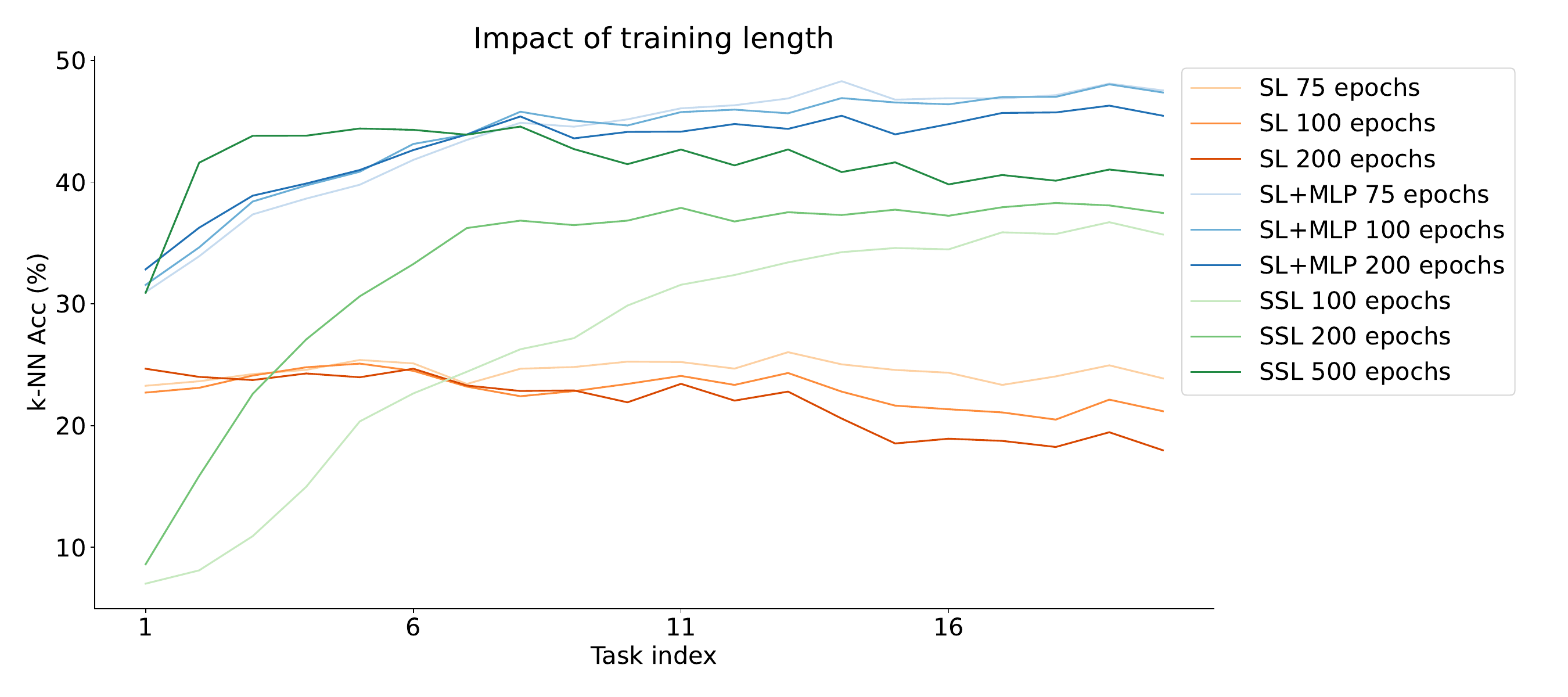}
    \caption{Self-supervised models benefit from longer training. However, supervised models, both with and without MLP projector result in reduced performance when trained for a large number of epochs. We report task-agnostic k-NN accuracy for all tasks after each task.
    } 
    \label{fig:epochs}
\end{figure}

We investigate how the number of epochs influences the representations trained with different methods. We conducted experiments on a long sequence of tasks, C100/20, training with SSL, SL, and SL+MLP methods for different numbers of epochs in each task. We present the results in Table~\ref{fig:epochs}. We observe that a large number of epochs (500) is important for SSL to achieve good final results. However, the performance gap between the SSL model trained on 500 epochs and the SSL models trained for 100 or 200 epochs is decreasing with a number of tasks.

\subsection{Task exclusion comparison}
\label{sec:apdx_task_excl_comp}

\begin{figure}[h!]
    \centering
    \includegraphics[width=0.85\textwidth]{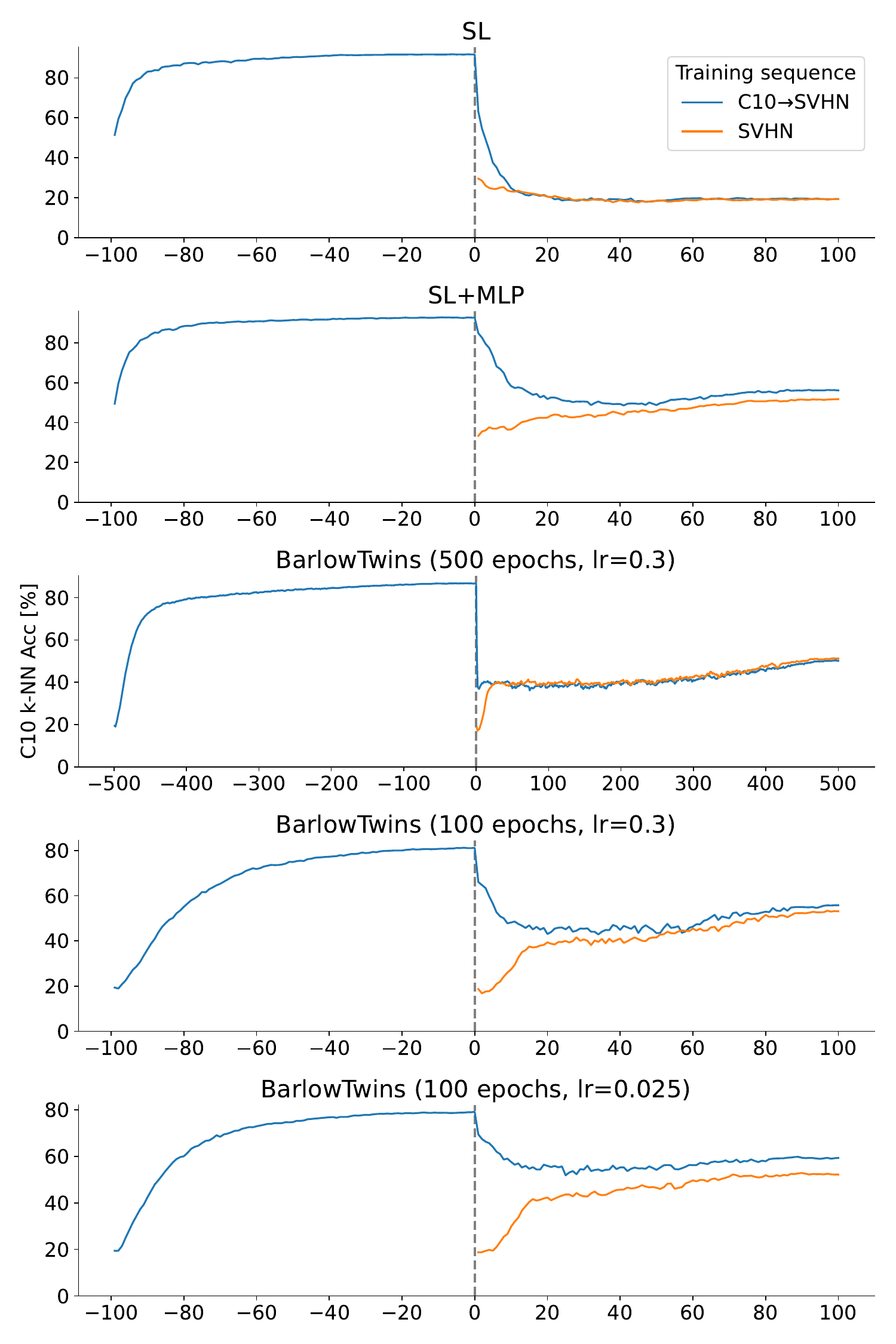}
    \caption{SSL behaves similarly to SL+MLP when trained for the same number of epochs with the same learning rate.} 
    \label{fig:feature-retention-extended}
\end{figure}

In Figure~\ref{fig:feature-retention-extended} we take a closer look at the task exclusion comparison. We identify that the training recipe is a factor responsible for its negative task exclusion difference. The training recipe for SL and SSL differs: SL is trained for 100 epochs with a 0.025 learning rate while SSL is trained for 500 epochs with 0.3 learning rate. When training SSL for 100 epochs with a learning rate of 0.025, following the SL+MLP learning recipe, we observe that SSL exhibits positive behavior that is similar to SL+MLP. However, such training configuration leads to the suboptimal final performance of a continual learner, as shown in Figure~\ref{fig:epochs}.

\subsection{Detailed two-task results}

\begin{figure}[h!]
    \centering
    \includegraphics[width=0.85\textwidth]{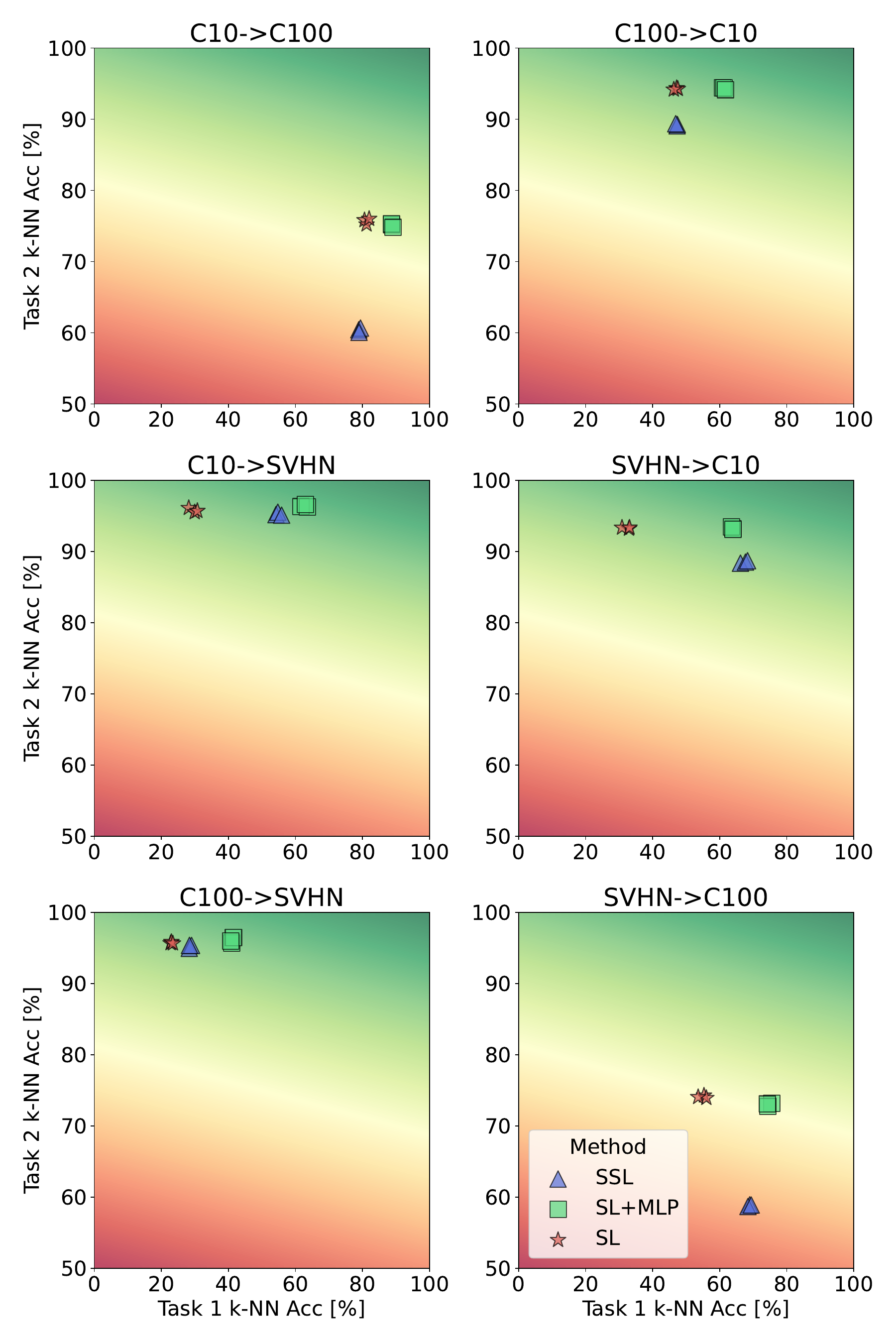}
    \caption{
        Results of two-task settings after training on the second task.
        Accuracy on the first task is presented on the horizontal axis and accuracy on the second task is presented on the vertical axis while the background color indicates the average accuracy on both tasks.
        SL usually outperforms SSL on the second task and usually underperforms on the first task. SL+MLP takes the best of both worlds (high first-task accuracy from SSL and high second-task accuracy from SL) and achieves the best overall performance.
    } 
    \label{fig:t0-vs-t1-acc}
\end{figure}

In Figure~\ref{fig:t0-vs-t1-acc} we present detailed results of two-task settings results summed up in Figure~\ref{fig:teaser}: C10$\xrightarrow{}$C100, C100$\xrightarrow{}$C10, C10$\xrightarrow{}$SVHN, SVHN$\xrightarrow{}$C10, C100$\xrightarrow{}$SVHN and SVHN$\xrightarrow{}$C100.
We can observe that self-supervised learning usually outperforms supervised learning on the first task. The opposite is true for the second task -- SL performs better than SSL. However, SL equipped with MLP achieves the highest average accuracy on both tasks usually outperforming both SL and SSL on the first and second tasks.

\end{document}